\pdfoutput=1
\documentclass[11pt]{article}

\usepackage[]{ACL2023}
\usepackage{times}
\usepackage{latexsym}
\usepackage[T1]{fontenc}
\usepackage[utf8]{inputenc}
\usepackage{microtype}
\usepackage{hyperref}
\usepackage{inconsolata}
\usepackage{booktabs}
\usepackage{inconsolata}
\usepackage{amsmath}
\usepackage{amssymb}
\usepackage{inconsolata}
\usepackage{graphicx}
\usepackage{subfigure}
\usepackage{caption}
\usepackage{subcaption}
\usepackage{booktabs}
\usepackage{multirow}
\usepackage{arydshln}
\usepackage{amsmath}
\usepackage{cases}
\usepackage{amsmath}
\usepackage{comment}
\usepackage{booktabs, multirow}
\usepackage{url}
\usepackage{pifont}
\usepackage{adjustbox}
\usepackage{soul}
\usepackage{rotating}

\usepackage{enumitem}

\newcommand{\method}{\textsc{CLeVer-CKE}}
\newcommand{\dataset}{\textsc{CroLin-MQuAKE}}
\newcommand{\mquake}{\textsc{MQuAKE}}
\newcommand{\pokemqa}{PokeMQA}
\newcommand{\mello}{MeLLo}
\newcommand{\pokemqacl}{PokeMQA-CL}
\newcommand{\mellocl}{MeLLo-CL}
\newcommand{\memit}{MEMIT}
\newcommand{\rome}{ROME}
\newcommand{\ft}{FT}

\newcommand{\llama}{LLaMa-2}
\newcommand{\vicuna}{Vicuna-1.5}
\newcommand{\chatgpt}{ChatGPT}

\definecolor{lightblue}{RGB}{173, 216, 230}
\definecolor{darkblue}{RGB}{0, 0, 139}

\title{Cross-Lingual Multi-Hop Knowledge Editing}

\author{{Aditi Khandelwal}\thanks{\enspace Equal Contribution (Randomly ordered via coin flip)} \textsuperscript{1,2}, {\bfseries Harman Singh\footnotemark[1] \thanks{\enspace Now at Google DeepMind}} , {\bfseries Hengrui Gu}\textsuperscript{4}, {\bfseries Tianlong Chen}\textsuperscript{3,5}, {\bfseries Kaixiong Zhou}\textsuperscript{4, 5}\\
        \textsuperscript{1}MILA - Quebec AI Institute, \textsuperscript{2}Microsoft, 
        \textsuperscript{3}UNC Chapel Hill, \\ \textsuperscript{4}North Carolina State University, \textsuperscript{5}Massachusetts Institute of Technology
        }

\begin{document}
\maketitle

\begin{abstract}
Large language models are often expected to constantly adapt to new sources of knowledge and knowledge editing techniques aim to efficiently patch the outdated model knowledge, with minimal modification. Most prior works focus on monolingual knowledge editing in English, even though new information can emerge in any language from any part of the world. We propose the \textit{Cross-Lingual Multi-Hop Knowledge Editing} paradigm, for measuring and analyzing the performance of various SoTA knowledge editing techniques in a cross-lingual setup. Specifically, we create a parallel cross-lingual benchmark, \dataset{} for measuring the knowledge editing capabilities. Our extensive analysis over various knowledge editing techniques uncover significant gaps in performance between the cross-lingual and English-centric setting. Following this, we propose a significantly improved system for cross-lingual multi-hop knowledge editing, \method{}. \method{} is based on a retrieve, verify and generate knowledge editing framework, where a retriever is formulated to recall edited facts and support an LLM to adhere to knowledge edits. We develop language-aware and hard-negative based contrastive objectives for improving the cross-lingual and fine-grained fact retrieval and verification process used in this framework. Extensive experiments on three LLMs, eight languages, and two datasets show \method{}'s significant gains of up to 30\% over prior methods. Code and data are released at \href{https://github.com/HarmanDotpy/CroLin-KE}{https://github.com/HarmanDotpy/CroLin-KE}
\end{abstract}
\section{Introduction}

\begin{figure}[t]
\centering
\includegraphics[width=\columnwidth]{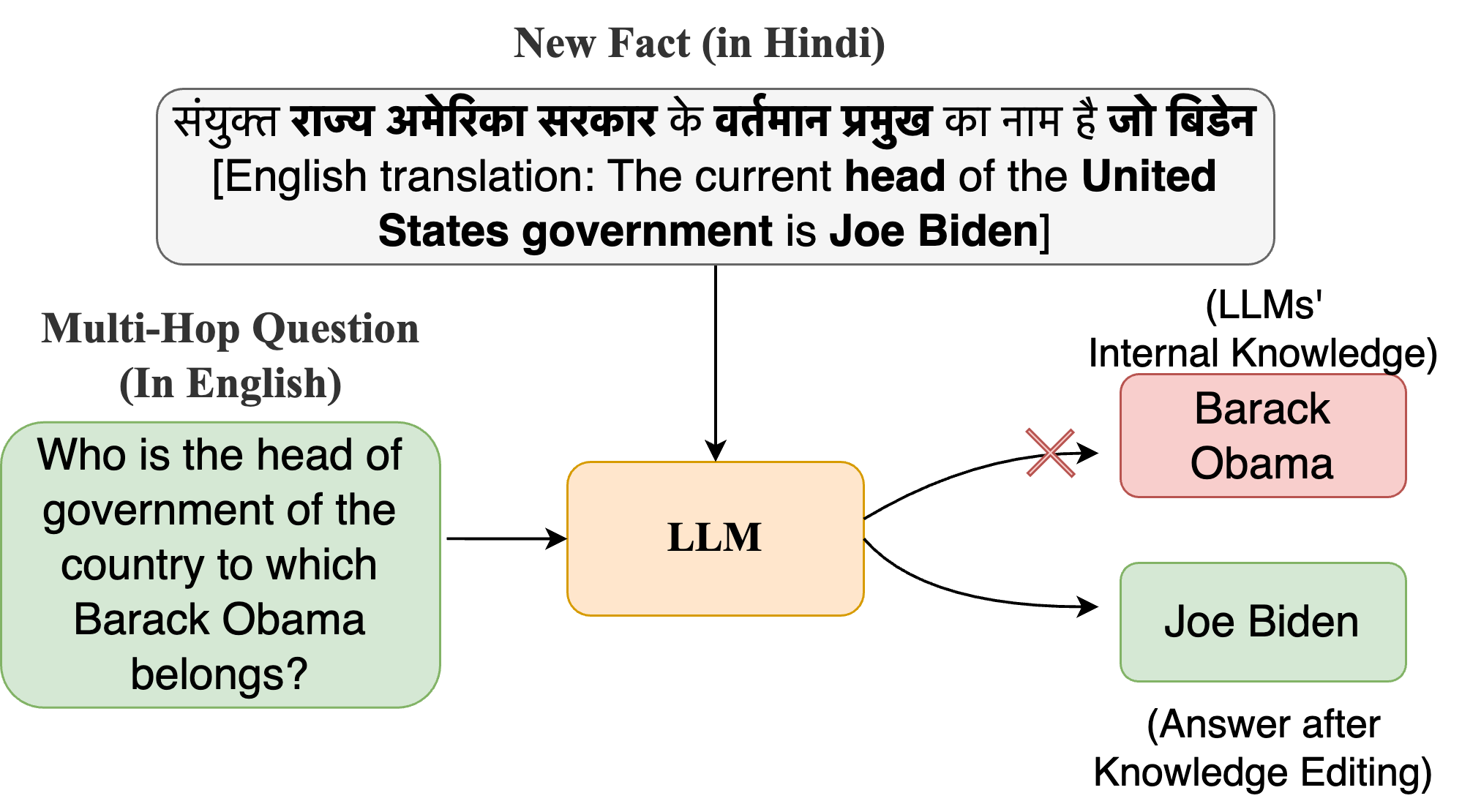}
\caption{The Cross-lingual Multi-hop knowledge editing problem. New fact(s) are provided in different languages (e.g. Hindi). An LLM should adapt to these facts for answering complex, multi-hop questions correctly in different languages (e.g. English).}
\label{fig:introfig}
\vspace{-0.3cm}
\end{figure}

Large language models (LLMs) are seeing an increasing adoption across users having different cultural and linguistic background, and need to be up to date about the ever-changing knowledge in the world for maintaining their utility and reliability in various applications. Due to the ever increasing compute and data requirements to train these models, there has been a surge in the development of knowledge editing techniques to modify the language models in an efficient way, such that they adhere to the world dynamics.

Prior work on knowledge editing has largely focused on editing LLMs in a monolingual setting \citep{zhong2023mquake, gu2024pokemqa}, where both 
user queries and edited facts are expressed in the form of English. These works can be grouped into two categories: parameter-update and parameter-preserving methods. The former directly updates the parameters within LLMs for updating knowledge about the edited facts through meta-learning, fine-tuning, or knowledge locating~\citep{de-cao-etal-2021-editing, knowledgeneurons, mend, ROME, memit}. The later approach freezes the parameters and explicitly stores the edited facts in an external memory and retrieves them for answering user queries~\citep{zhong2023mquake, gu2024pokemqa, serac, hartvigsen2023aging}. Existing monolingual knowledge editing techniques aren't broadly applicable since new knowledge can emerge in different languages. Some works have made progress in this direction \citep{beniwal2024crosslingual, xu-etal-2023-language-anisotropic, si2024mpn}, but they have considered a simplistic setting of assuming the edited facts as independent without any multi-hop rippling consequences on entailed reasoning process, and are primarily focused on parameter-modifying based editing methods.

There has only been a limited focus on the realistic case of \textit{cross-lingual multi-hop knowledge editing} (see Fig~\ref{fig:introfig}), where the edited knowledge can come in through users who communicate in different languages. Further, much of edited knowledge often has a rippling effect on other facts of the world. For example, the \textit{club change of Messi} affects deduction process of question "indicating a superficial word matching rather than a contextual grasp of the entities involved." This knowledge editing setting, which we argue is important to study, is challenging since the model needs to transfer knowledge about fact edits between different languages, while also reasoning about the facts which are modified as a consequence to the given edit. Poor knowledge transfer between languages can lead to error propagation across reasoning steps which can increase failure cases of model editing.

In this work, we formulate the notion of cross-lingual multi-hop knowledge-editing and analyze existing approaches for their editing ability in different languages, following which a simple yet highly effective approach is designed. Specifically,\\
\textcolor{blue}{\ding{172}} We create one of the first benchmark datasets for measuring cross-lingual multi-hop knowledge editing capabilities of knowledge editing methods. Besides parameter-update based approaches, we contribute strong retrieval-based baselines for knowledge editing and provide a comprehensive analysis.\\
\textcolor{blue}{\ding{173}} We provide a detailed analysis and find significant gaps in the performance of methods for cross-lingual knowledge editing. The gaps are mainly due to challenges in accurately recalling fact edits made in language other than input query.\\
\textcolor{blue}{\ding{174}} To bridge such gap, we design a competitive method, termed as \underline{C}ontrastive \underline{L}anguag\underline{e}-aware \underline{Ver}ification for \underline{C}ross-lingual \underline{K}nowledge \underline{E}diting (\method{}), for improving performance of cross-lingual multi-hop knowledge editing. Our approach is based on decomposing a multi-hop question in a particular language into sub-questions and retrieving fact edits (if any) from memory using a cross-lingual retriever, which is integrated for answering sub-questions. In particular, the cross-lingual retriever is regularized by novel language-guided and hard-negative based contrastive losses, which leads to improved language and fine-grained sentence understanding of the edits, leading to high quality cross-lingual retrievals. \method{} improves over previous SoTA by up-to 30\% increase in knowledge editing accuracy when tested on multiple LLMs, datasets and languages.

\section{Cross-lingual Multi-hop Editing}
Following prior work \cite{zhong2023mquake}, a fact is defined as a triplet $(s,r,o)$, where $s$ is the subject, $o$ is the object, and $r$ is the relation (e.g., \textit{Shakespeare, author of, Hamlet}). Given that a parametric LLM can become outdated or incorrect, knowledge editing is required to be performed on it. 
An edited fact stores information about updated knowledge of an existing fact and is denoted as 
$e = (s,r,o^*)$, where the object is replaced with a new one $o*$.

\noindent\textbf{Cross-Lingual Knowledge Editing.}
Each knowledge fact or edit is assumed to be represented in natural language. Let $\mathcal{T}: \mathcal{E} \rightarrow \mathcal{L}$ be a function which takes any fact $e \in \mathcal{E}$ (e.g., \textit{Shakespeare, author of, Hamlet}) and converts it into a natural language statement, (e.g., \textit{Shakespeare is the author of Hamlet}). All the facts and edits can be represented in a variety of languages $\{L_1, L_2, \dots\}$ 
via functions such as $\{\mathcal{T}_{L_1}, \mathcal{T}_{L_2}, \dots\}$. 
For example, an edit $e = $\textit(Shakespeare, author of, Lolita) can be written as $\mathcal{T}_{\mathrm{de}}(e)=$ \textit{Shakespeare ist der Autor von Lolita} in German and $\mathcal{T}_{\mathrm{en}}(e)=$ \textit{Shakespeare is the author of Lolita} in English.

We consider a  collection of $n$ fact edits in the diverse languages: $\mathcal{E} = \{e_1^{L_1}, e_2^{L_2}, e_3^{L_2}, ..., e_n^{L_i}\}$, where $L_1, L_2, ..., L_i$ are different languages for e.g., German, Hindi, Swahili, etc. A language model $f$ is said to be edited with  new knowledge facts if the model generations adheres to all the edits present in $\mathcal{E}$. The model is required to seamlessly transfer knowledge about an edit 
in one language to answer queries in other languages.

\noindent\textbf{Multi-Hop Editing and Evaluation.}
We follow \citet{zhong2023mquake} for evaluating knowledge editing via multi-hop question answering. Consider $e_{L_1} = ( s_{i}^{L_1}, r_{i}^{L_1}, o_{i}^{L_1*})$, an edited fact in language $L_1$. Also consider a chain of facts $\mathcal{P} = \langle (s_{1}^{L_1},r_{1}^{L_1},o_{1}^{L_1}), \dots , (s_{n}^{L_k},r_{n}^{L_k},o_{n}^{L_k}) \rangle$, where object of a fact is the subject for the next fact. Any edit to the first fact $(s_{1}^{L_1},r_{1}^{L_1},o_{1}^{{L_1}*})$ will likely have a rippling effect and change the subsequent facts in the chain, and we expect a successfully edited model to be aware of all such entailed changes.

For evaluating models in a cross-lingual multi-hop setting, we make use of multi-hop questions such as $Q_{L_n}$, in language $L_n$ which is different from $L_{1\dots k}$. The question asks about the head entity $s_{1}^{L_1}$ for which the answer is $o_{n}^{L_k}$ before editing. After editing, the fact chain changes to $\mathcal{P^*} = \langle (s_{1}^{L_1},r_{1}^{L_1},o_{1}^{{L_1}*})$ $,(s_{2}^{L_2},r_{2}^{L_2},o_{2}^{{L_2}*}), \dots , (s_{n}^{L_k},r_{n}^{L_k},o_{n}^{{L_k}*}) \rangle$ since edits in the first fact can affect the subsequent facts it's linked to.
For answering $Q_{L_n}$ after editing, the model has to account for this rippling effect, and provide the final answer as $o_{n}^{{L_k}*}$. For this, model has to transfer knowledge of the edited fact and the answer, between languages $L_{1\dots k}$ and $L_n$, while correctly reasoning about fact edits via $\mathcal{P^*}$.
\section{\dataset{} Benchmark}
\label{s:dataset}

We develop one of the first parallel cross-lingual and multi-hop benchmarks 
for measuring the knowledge editing capabilities of the existing approaches. A parallel benchmark across languages has the same test examples across all the languages, enabling a direct comparison between them. For this, we use existing datasets measuring the multi-hop model editing in English: MQuAKE-CF and MQuAKE-T released by \citet{zhong2023mquake}, which have counterfactual edits and real-world temporal edits respectively. We translate one fact edit per example in these datasets using Google Translate \citep{GoogleTranslate} into 7 languages with diverse writing scripts across medium to high resourcedness - German, Spanish, Chinese, Russian, Hindi, Bengali, Swahili. This results in the benchmark: Cross-Lingual Multi-Hop QnA for Knowledge Editing (\dataset{}). It has two datasets, \dataset{}-CF and \dataset{}-T, each having 8 languages, and 3k and 1.8k parallel examples (same examples in all languages) per language, respectively. The translations are verified by human experts proficient in particular languages and evaluation of BLEU score \cite{papineni-etal-2002-bleu} using backtranslation. We find that the translation is highly accurate, since we study medium to high resource languages. See Section \ref{app:translation_verif} for more details. 

Concurrently, \citet{wei2024mlake} created a multilingual knowledge editing dataset using Wikipedia, offering translocalized knowledge but lacking parallel multilingual examples like ours. \dataset{} enables comparing the knowledge editing performance difference across languages directly without being affected by the variation of test sets between different languages.
\section{Benchmark Analysis on Cross-Lingual Multi-hop Knowledge Editing}
\label{expt}

\begin{table*}[t]
\centering
\resizebox{0.95\textwidth}{!}
{
    \footnotesize
    \begin{tabular}{lcccccccccc}
        \toprule
        \multicolumn{1}{c}{} & & \multicolumn{4}{c}{\dataset{}-CF} & & \multicolumn{4}{c}{\dataset{}-T} \\
        \cmidrule(lr){3-6} \cmidrule(lr){8-11}
        \multicolumn{1}{c}{} & & \multicolumn{2}{c}{3k (All)} & \multicolumn{2}{c}{100 edited} & & \multicolumn{2}{c}{1.8k (ALL)} & \multicolumn{2}{c}{100 edited} \\
        \cmidrule(lr){3-4} \cmidrule(lr){5-6} \cmidrule(lr){8-9} \cmidrule(lr){10-11}
        \multicolumn{1}{l}{Method} & & Acc. & Hop-Acc & Acc. & Hop-Acc & & Acc. & Hop-Acc & Acc. & Hop-Acc \\
        \midrule
        \multicolumn{9}{c}{\textbf{\llama{}}} & \multicolumn{2}{r}{Size: 7B} \\
        \midrule
        
        \ft{} & & 0.0  & 0.0 & 0.3 &0.0 & & 0.0 & 0.0 & 0.0 & 0.0 \\
        \rome{} & &  1.9 & 0.0 & 2.3 & 0.0 & & - & - & - & -\\
        \memit{} & &  0.4 & 0.3 & 4.2 & 1.0 & & - & - & - & - \\

        \mellocl{}  & &  10.6 & 1.9 & 14.6 & 2.3    & &  26.5 & 3.0 & 28.5 & 0.7  \\
        \pokemqacl{} 	& 	& 10.6 & 2.3 & \textbf{19.7} & 5.9 & & 11.1 & 5.8 & 14.6 & 7.8\\
        \method{}    &    &  \textbf{13.2} & \textbf{7.3}      &  19.2 & \textbf{11.1} & & \textbf{40.6} & \textbf{30.0} & \textbf{42.6} & \textbf{31.1} \\
        \midrule
        \multicolumn{9}{c}{\textbf{\vicuna{}}} & \multicolumn{2}{r}{Size: 7B} \\
        \midrule
        \mellocl{}  &  & 8.8 & 2.8 & 14.5 & 5.5 &  & 34.1 & 13.5 & 36.9 & 13.0 \\
        \pokemqacl{}  &  & 9.5 & 2.1 & 17.3 & 5.5 & &  11.0 & 6.6 & 13.7 & 8.5 \\
        \method{}  &  & \textbf{12.7} & \textbf{7.1} & \textbf{18.1} & \textbf{10.7} &  & \textbf{ 37.9} & \textbf{30.6} & \textbf{39.9} & \textbf{31.8}  \\
        \midrule
        \multicolumn{9}{c}{\textbf{\chatgpt{} (GPT-3.5-turbo-instruct)}} & \multicolumn{2}{r}{Size: Undisclosed} \\
        \midrule
        \mellocl{}  &  & 14.4 & 5.4 &  20.6 & 8.5  & &  39.0 & 17.6 & 41.4 & 17.0 \\
        \pokemqacl{} &  & 12.9 & 2.9 & 26.8 & 9.3 & & 13.5 & 8.2 & 17.4 & 10.7  \\
        \method{}  &   & \textbf{18.6} & \textbf{10.6} & \textbf{30.1} & \textbf{18.6} & & \textbf{42.6} & \textbf{32.8} & \textbf{45.6} & \textbf{35.1} \\
        \bottomrule
    \end{tabular}
}
\caption{Performance of parameter update based and in-context editing based methods on the cross-lingual multi-hop knowledge editing problem, reported for three language models, and averaged over 8 diverse languages. Parameter-update based methods -- \ft{}, \rome{}, \memit{} perform significantly worse than in-context editing methods, \mellocl{}, \pokemqacl{} and \method{}, significantly outperform all baselines. Evaluation is performed over two sizes of edited fact memory -- 100 and 3k/1.8k following \citet{zhong2023mquake}. See \S\ref{expt} for more details.}
\label{t:multilinugal_avg_mainresults}
\end{table*}

\noindent\textbf{LLMs.} We use SoTA propriety and open-source LLMs: \chatgpt{} \cite{chatgpt}, \llama-7B \cite{touvron2023llama}, \vicuna-7B \cite{vicuna2023} as backbones to evaluate cross-lingual multi-hop knowledge editing.


\noindent\textbf{Evaluation Metrics.} We use  multi-hop accuracy proposed by \citet{zhong2023mquake} which measures the accuracy of the final answer of a multi-hop question. We also adopt hop-wise answering accuracy for checking the correctness of intermediate reasoning steps, as proposed by \citet{gu2024pokemqa}.

\noindent\textbf{New Baselines.} Based on existing work, we contribute strong baselines for the new editing setup:
\begin{itemize}[noitemsep,nolistsep,leftmargin=*]
    \item \textbf{\mellocl{}:} We modify the existing method of \mello{} \citep{zhong2023mquake} by replacing the monolingual retriever used in their system 
    with a multilingual retriever.  
    This minimal modification allows the system to retrieve the cross-lingual edits. 
    \mellocl{} is a simple retrieval-based knowledge editing approach: LLM first breaks down a multi-hop question into various sub-questions and for each sub-question, the retriever then recalls the most relevant fact from an external memory. The LLM disambiguates if the retrieved fact is useful for answering the question or not.
    \item \textbf{\pokemqacl{}:} \pokemqa{} \citep{gu2024pokemqa} is similar to \mello{} but consists of a conflict disambiguator for retrieving as well as classifying if a fact is useful to answer a sub-question. Following \pokemqa{}, we train this disambiguator using BCE loss with negative sampling for retrieving the close edits, given a decomposed sub-question. 
    However, our training dataset now consists of translated version of the training dataset used in \pokemqa{}. This training set contains all 8 languages (the multilingual setting) or English along with one of the 7 non-English languages (the bilingual setting). 
\end{itemize}

\begin{figure}[h] 
  \centering
  \includegraphics[width=0.5\textwidth]{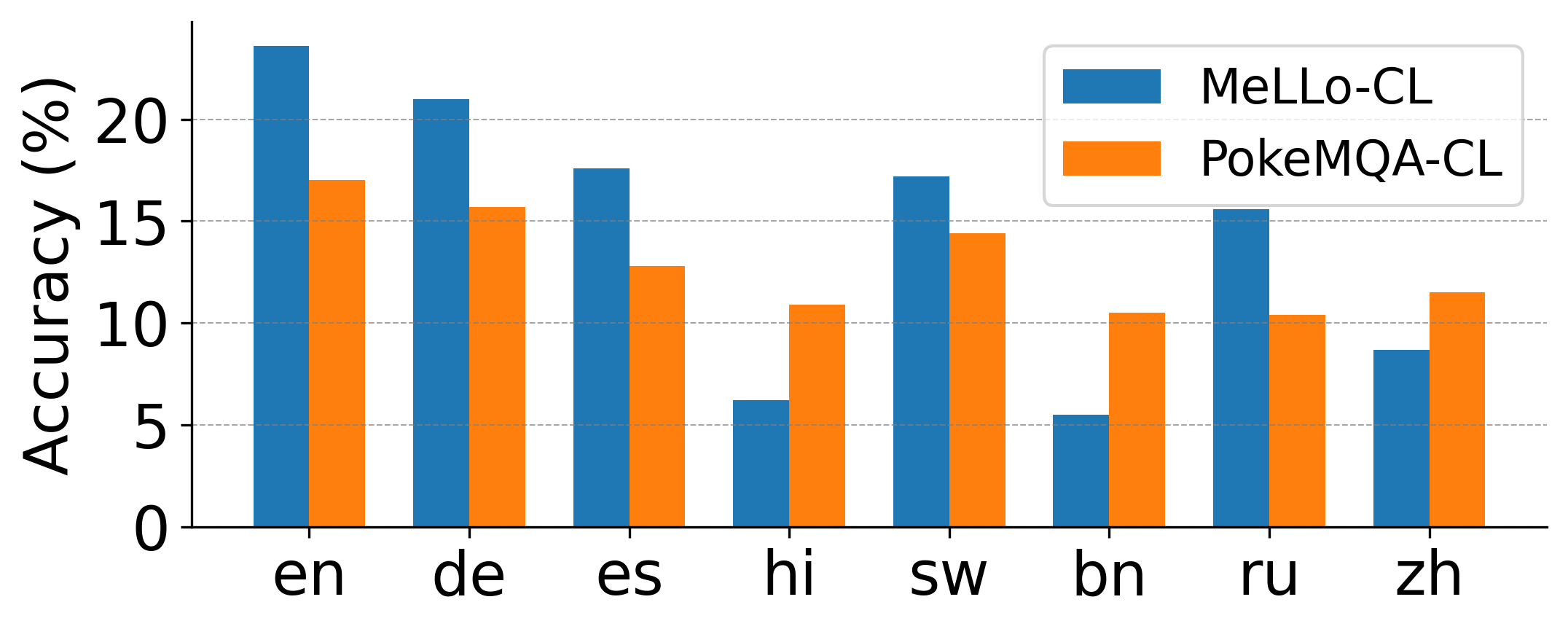} 
  \caption{Comparison of multi-hop accuracy of Mello-CL and \pokemqa{}-CL on the \dataset{}-CF across the different languages.}
\label{fig:mello_pokemqa_multilingual}
\end{figure}

\noindent\textbf{Multi-hop knowledge editing performance heavily depends on the language of edits.} As can be seen in the Figure~\ref{fig:mello_pokemqa_multilingual}, the gaps in average accuracy between English and other language edits are 10\% and 11.7\% for methods \mellocl{} and \pokemqacl{}, respectively,  highlighting the significant drop in cross-lingual knowledge editing setup. Performance of \mellocl{} varies significantly across the different scripts. For language written in Latin scripts, the accuracy is $\sim$20\%. In contrast, for languages written in non-Latin scripts such as Devanagari, Chinese, or Cyrillic, the accuracy drops to $\sim$11\%. Another observation is that, in case of edits made in Swahili, despite being a low-resource language, it outperforms more resource-rich languages like Chinese, Russian, and Hindi. This suggests that script plays a crucial role in cross-lingual knowledge editing and retrieval. The reason is intuitive, i.e., Latin script languages have a higher presence in most pretraining data which leads to better tokenization and better representation in LLMs; whereas the non-Latin script languages suffer from high tokenization fertility and less effective representation in the model \citep{Ahia2023DoAL, singh-etal-2024-indicgenbench}.

\noindent\textbf{Parameter-modifying based knowledge editing performs poorly in the cross-lingual setting.}
Methods that update the parameters of the model, such as \rome{}, \memit{}, \ft{}, perform significantly worse in the cross-lingual setting, achieving an accuracy under 5.0\% (average across languages), as shown in Table~\ref{t:multilinugal_avg_mainresults}. One key issue is that knowledge edits may not transfer effectively across different languages just via model weights, leading to inconsistent and inaccurate retrievals. Further, the problem is exacerbated due to cascading error propagation in a multi-hop setting. Hence the parameter-modifying methods struggle to reliably edit the LLM across languages and multi-hop contexts. This highlights the need for memory-based approaches that rely on an external edit memory, like our contributed baselines, \mellocl{} and \pokemqacl{}, which can cross-lingually retrieve the relevant edits on the fly when inferring from an LLM. These approaches substantially improve performance up to nearly 30\% on \dataset{} compared to parameter-modifying based methods.

\noindent\textbf{Knowledge editing performance based on retriever training technique.} 
\mellocl{} retrieves the edited fact from the memory using mContriever and employs an LLM to disambiguate between the generated answer and the retrieved fact and hence ascertains if the generated fact needs any update or not. On the other hand, the current state-of-the-art knowledge editing method in English, \pokemqacl{}, uses a retrieve-then-verify approach, which offloads the knowledge disambiguation to the retriever. This retriever is a light-weight and fine-tuned distilbert-base model trained on a (sub-question,edit) pair dataset using binary cross-entropy loss with negative sampling. It retrieves the closest edits (in fact memory) to a sub-question and scores it on whether the edit answers the question or not (called verification or disambiguation). If it does, then it uses this new knowledge as the answer to the sub-question in the \textit{n}-th hop step and performs in-context editing. \pokemqacl{} outperforms \mellocl{} on in the monolingual (English) setting, with a much smaller retriever as shown in \citet{gu2024pokemqa}, however, when trained with multilingual data, we find that it \textit{significantly under-performs \mellocl{}} in most languages including English as shown in Fig. \ref{fig:mello_pokemqa_multilingual}. \mellocl{} under-performs in Hindi and Bengali -- languages with scripts very different from Latin, even though its retriever is trained with 100+ languages.

\noindent\textbf{Qualitative analysis of errors.}
\label{p:error_analysis}
We examine the error cases of \mellocl{} and \pokemqacl{} for knowledge edits made in two languages: English and Hindi. Our analysis identifies two primary types of errors made by these methods. The first type is \underline{a) incorrect retrieval}, where the retrieved information is not relevant to input queries. The second type is \underline{b) incorrect LLM response}, where a LLM either makes a mistake in extracting the final answer or errors in decomposing the question into subquestions. Additionally, \mellocl{} exhibits \underline{c) contradiction error} where the LLM makes mistake at the contradiction step. Figure~\ref{fig:error_types} illustrates the examples of these three types of errors.
We analyzed a random subset of 30 samples for these methods and found the following: 

\ding{182} \mellocl{}: When edits are made in English, 63.3\% of the samples are correct, 29.3\% have the {contradiction error}, 3.6\% have {Incorrect retrieval}, and 3.6\% have the {incorrect LLM response}.
For edits made in Hindi, 33.3\% of the samples are correct, 60\% exhibit an error combination of {incorrect retrieval} and subsequent {contradiction error}, where the model first makes an incorrect retrieval and then fails in the contradiction step and 6.6\% of erroneous samples are due to the {incorrect LLM response}. In the \dataset{}-CF case when the multilingual edited fact memory contains edits in English and Hindi, \mellocl{}'s retriever rarely retrieves edits in Hindi, indicating a limitation in its multilingual capabilities. The limitation of \mellocl{} lies in its retriever-then-contradict mechanism which is up to the LLM. 

\ding{183} \pokemqacl{}: When edits are made in English, 53.3\% of the samples are correct and 46.3\% have the incorrect retrieval error. When edits are made in Hindi, 43.3\% are correct, 51\% have errors due to the {incorrect retrieval} and 5.6\% are due to the {incorrect LLM response}. The limitation of \pokemqacl{} lies in its reliance on a bag-of-words model for retrieval. For instance, when presented with the sub-question "\textit{Who is the \textbf{head of state} of the USA?}", it retrieves the fact ``\textit{The \textbf{head of state} of Mongolia is Khürelsükh Ukhnaa.}" This example underscores that \pokemqacl{} prioritizes facts with the highest word overlap, specifically ``\textit{head of state}" indicating a superficial word matching rather than a contextual grasp of the entities involved. 

\ding{184} When trained in a cross-lingual setting, \pokemqacl{} exacerbates the issue of bag-of-words retrieval. For example, for the sub-question ``\textit{Where was \textbf{Bob Dylan} born?}", it correctly retrieves ``\textit{\textbf{Bob Dylan} was born in the city of Nankoku}" in English. However, if the same edit is made in German, it retrieves ``\textit{\textbf{Bob Dylan} spricht die Sprache von Malayalam}" (\textbf{Bob Dylan} speaks the language of Malayalam). This issue is a likely a consequence of high word overlap in retriever's internal translation process and is a limitation of current systems.

Section \ref{expt} hints signficant gapS between English-only and cross-lingual case, and that proper knowledge retrieval technique is critical to the performance of cross-lingual knowledge editing.
\section{\method{} for Knowledge Editing}
\label{method}

\begin{figure*}[t]
\centering
\includegraphics[width=0.9\textwidth]{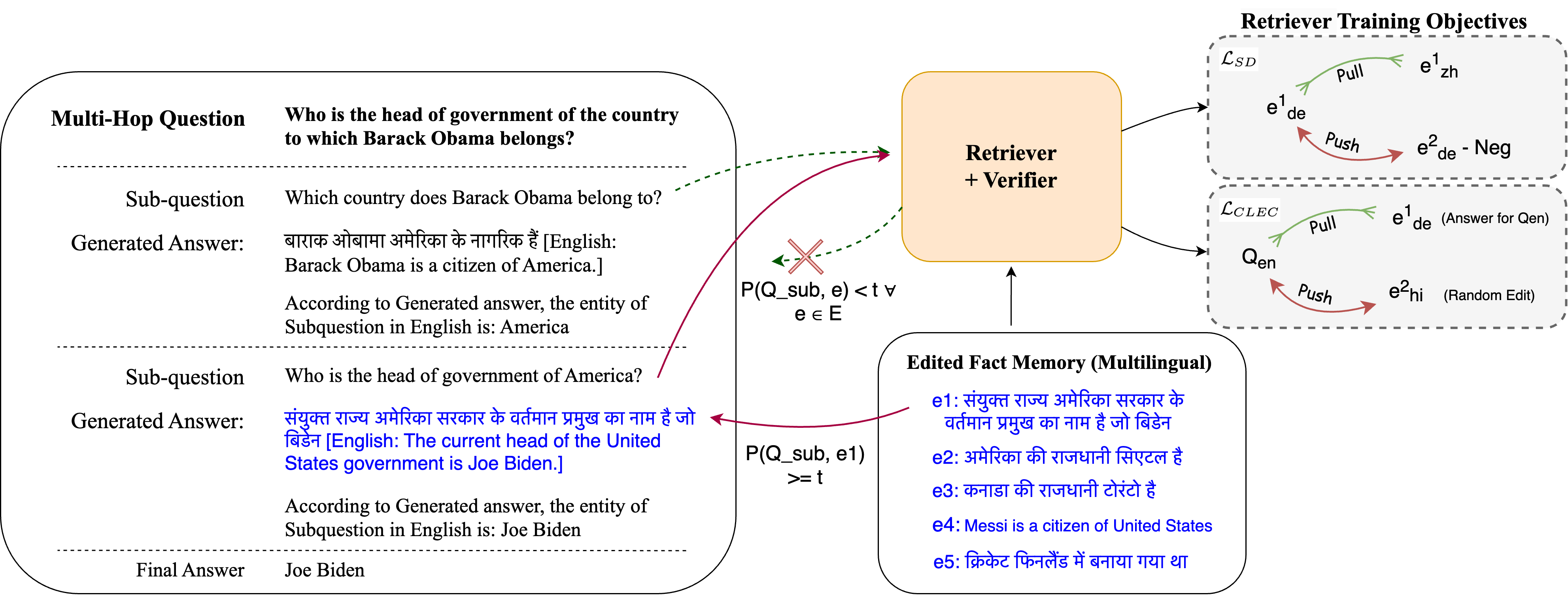}
\caption{Our proposed method, \method{}. On the left we show the LLM inference process for cross-lingual multi-hop knowledge editing. Given a prompt (See \S\ref{prompts}), the LLM breaks down a multi-hop question into sub-questions and answers them individually, utilizing a a retrieve and verify approach using the retriever. On the right, we show new training objectives used in this work for training the retriever. See \S\ref{method} for more details.}
\label{fig:multi_hop_cross_ling}
\vspace{-0.3cm}
\end{figure*}

For overcoming limitations in cross-lingual multi-hop knowledge editing, we design \method{}, a cross-lingual and light-weight model editor that seamlessly integrates into any backbone LLM, without changing its parameters.
\method{} is inspired by memory-based and retrieval-augmented knowledge editing methods \citep{zhong2023mquake, gu2024pokemqa, mitchell2022fast} for mutlihop question answering. \method{} follows the following procedure: Given an input query, it \textbf{a)} decomposes the multi-hop question into multiple sub-questions for getting to the final answer, and for answering each sub-question \textbf{b)} retrieves a relevant fact from the edit memory, \textbf{c)} disambiguates whether the retrieved new knowledge is relevant to answering the sub-question, and \textbf{d)} continues the model generation process based on that. In this work, we primarily aim at showing the importance of having a high-quality retriever for the retrieve-and-verify steps at \textbf{b)} and \textbf{c)} described as follows. See Fig. \ref{fig:multi_hop_cross_ling} for an overview.

\paragraph{Memory of Fact Edits:} \method{} explicitly stores a set of knowledge edits $\mathcal{E}$ in a memory $\mathcal{F}$. Each edit triplet $e=(s,r,o)\in\mathcal{E}$ is converted to a natural language statement in either English or another language using English or translated templates present in \dataset{}. This creates a multilingual edited fact memory.

\paragraph{Sub-question Decomposition:} Given a multi-hop question $Q$, LLM is prompted using in-context examples to decompose it into various sub-questions $Q_{\mathrm{sub}} = \{q_1, q_2, \dots\}$. Note that $Q$ and the language model generation is assumed to be in English in our work whereas the edited fact memory can contain both English and non-English knowledge edits. The LLM is instructed to answer the generated sub-questions as follows.

\paragraph{Retrieve-and-Verify:} For each sub-question $q$, \method{} retrieves the top-1 candidate $r \in \mathcal{F}$ using cosine similarity. Verification process then answers the question: \textit{Does $r$ help answer $q$?} The answer to this is yes if $cos(f(r), f(q))\geq t$ where $cos(.)$ is the cosine similarity function, $f(.) \in \mathbb{R}^{d}$ is the retriever embedding and $t$ is a threshold (hyperparameter). In this case, $r$ is passed to the LLM which uses it for generating the answer to the sub-question. If $cos(f(r), f(q))<t$, only the LLM's internal knowledge is used to answer the question. Following this, LLM will move on to answering the next sub-question. Note that here, the disambiguation of whether $r$ is useful or not, happens external to the LLM, reducing its reasoning complexity.

\paragraph{\method{} Retriever Training:} 
Motivated by gaps found in Section \ref{expt}, we create new objectives for training the retriever for improving fine-grained and cross-lingual representations. We then show that our simple losses provide significant gains in knowledge editing performance.

\underline{Semantic Distinction Loss:} We employ a contrastive, triplet margin loss $\mathcal{L}_{\mathrm{SD}}$ for improving fine-grained cross-lingual retrieval. Assuming an edits $e=(s,r,o)$, we obtain its natural language forms $\mathcal{T}_{L_1}(e)$, $\mathcal{T}_{L_2}(e)$ in languages $L_1$, $L_2$ respectively. This creates a positive pair for the triplet loss. We generate hard negatives for $\mathcal{T}_{\mathrm{en}}(e)$ in English by replacing an edits' subject, object, or both object with random entities, with a probability of 0.33 each. This process involves extracting all relations in \mquake{} dataset and prompting the GPT-3.5 model to suggest head/tail entities for these relations. We then randomly sample any generated head/tail (or both) for replacement in an edit containing the corresponding relation. Following this, the hard negative example $\mathcal{T}_{\mathrm{en}}(e_\mathrm{neg})$ is translated to $L_1$ and hence a negative pair $(\mathcal{T}_{L_1}(e), \mathcal{T}_{L_1}(e_\mathrm{neg})$ is obtained. The loss function is formulated as:
\begin{equation}
\small
\begin{split}
\mathcal{L}_{\mathrm{SD}} = \max(& d(f(\mathcal{T}_{L_1}(e)), f(\mathcal{T}_{L_2}(e)) \\
 - & d(f(\mathcal{T}_{L_1}(e)), f(\mathcal{T}_{L_1}(e_{\mathrm{neg}})) + \alpha, 0).
\end{split}
\end{equation}
$f(\cdot)$ represents the retriever embedding, $d(.)$ represents the distance function, and $\alpha$ is a gate hyperparameter. $\mathcal{L}_{\mathrm{SD}}$ promotes learning the fine-grained knowledge about subject, relation and object in a cross-lingual setting and encourages the model to distinguish the semantic nuances in different edits. This mitigates the redundant selection of edits with significant word overlap.

\underline{Cross-Lingual Edit Consistency Loss:} We employ a contrastive, triplet margin loss $\mathcal{L}_{\mathrm{CLEC}}$ focused on improving cross-lingual retrieval. Here, the anchor is $Q_{\mathrm{en}}$, a question in English. The edited fact for answering that question, $\mathcal{T}_{L_1}(e)$, serves as the positive example, and a random edit $\mathcal{T}_{L_2}(e_{\mathrm{rand}})$ forms the negative example:
\begin{equation}
\small
\begin{split}
\mathcal{L}_{\mathrm{CLEC}} = \max(& d(f(Q_{\mathrm{en}}), f(\mathcal{T}_{L_1}(e)) \\
 - & d(f(Q_{\mathrm{en}}), f(\mathcal{T}_{L_2}(e_{rand})) + \alpha, 0).
\end{split}
\end{equation}

\underline{BCE Loss:} Following \citep{gu2024pokemqa, NIPS2013_9aa42b31} we add a binary cross-entropy loss in the cross-lingual setting as a baseline loss for training the retriever for retrieving edits in a cross-lingual setting.  The negative BCE Loss function takes questions in English and their corresponding edited facts in one of the seven languages as input. We then compute the $L_2$ norm between these edits and questions, and sample 20 negatives. The loss function \(\mathcal{L}\) is defined similar to \citet{gu2024pokemqa}:
\begin{equation}
\small
\begin{split}
\mathcal{L}_{\mathrm{BCE}} = - & \log g(\mathcal{T}_{L_1}(e), f(Q_{\mathrm{en}})) \\
- & \mathbb{E}_{q_n \sim P_n(q)} [\log(1 - g(\mathcal{T}_{L_1}(e), q_n))],
\end{split}
\end{equation}
where \(P_n\) is a uniform over each mini-batch, and $g(.) = exp(d(.))$.

$\mathcal{L}_{\mathrm{CLEC}}$ and $\mathcal{L}_{\mathrm{BCE}}$ encourage it to differentiate between edits in different languages and enhance its ability to handle multilingual knowledge editing tasks effectively. The total loss we use is then:
\begin{equation}
\mathcal{L}_{\mathrm{total}} = \mathcal{L}_{\mathrm{SD}} + \mathcal{L}_{\mathrm{CLEC}} + \mathcal{L}_{\mathrm{BCE}}.
\end{equation}
\subsection{Performance of \method{}}

\begin{figure}[h]
  \centering
    \includegraphics[width=\columnwidth]{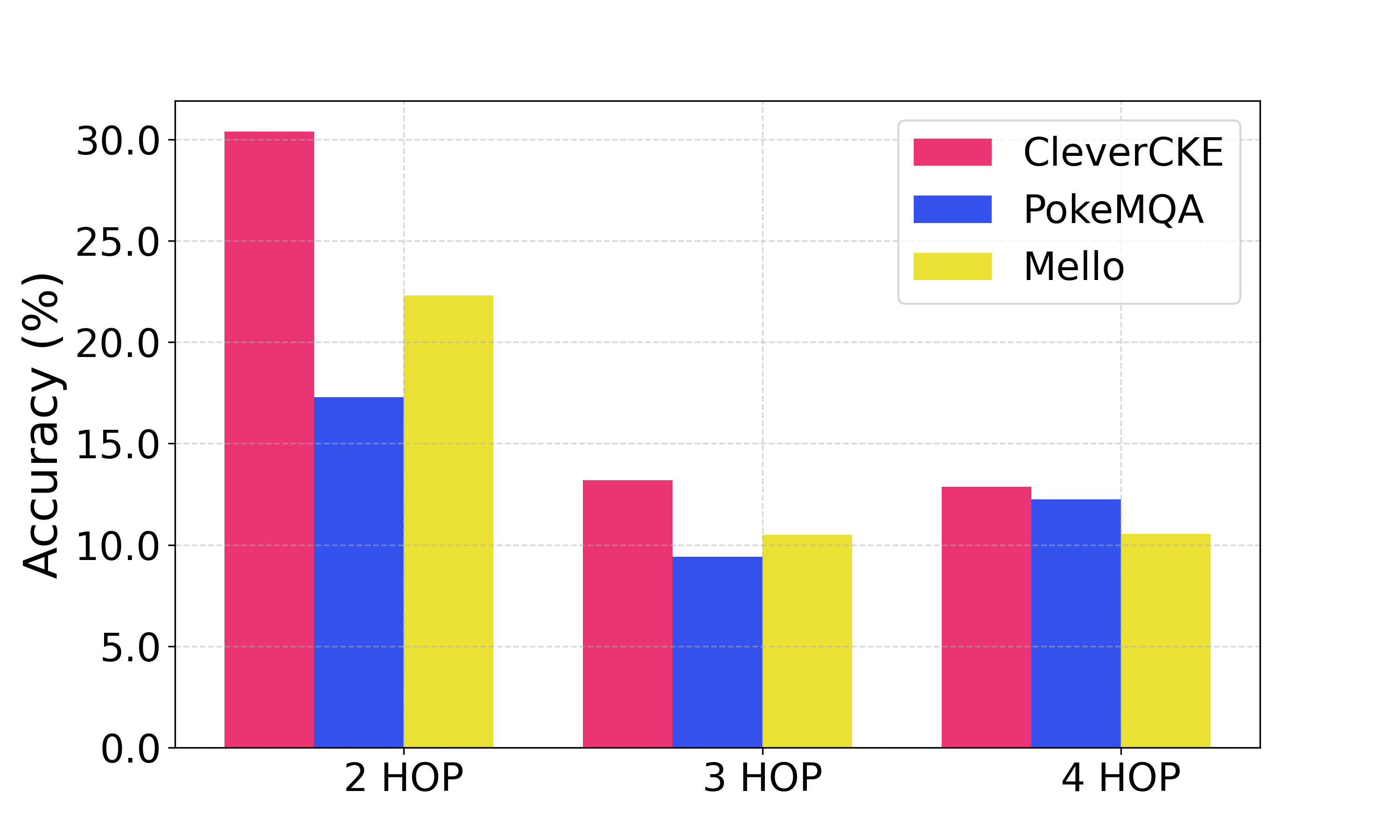}
    \caption{Average accuracy of methods \method{}, \pokemqa{}-CL and MeLLo-CL reported on 2, 3, 4-hop questions with ChatGPT as LLM with the case of all edited on \dataset{}-CF.}
    \label{fig:hop_ablation_avg}
    \vspace{-0.3cm}
\end{figure}

We train the retriever with the above losses on a dataset of 8 languages and measure performance on the \dataset{}. In Table~\ref{t:multilinugal_avg_mainresults}, on average across languages and across different LLMs, \method{} improves over previous methods by up-to 5.7\% in accuracy on \dataset{}-CF and we see a much larger increase in the hop-accuracy which suggests faithful reasoning. On the real world temporal dataset \dataset{}-T, we see a significant increase of about 30\% accuracy and more than 25\% in hop-accuracy metric. Performance gains are large and consistent or better for larger and more capable models like \chatgpt{}, as compared to \llama{}/\vicuna{}. Refer to Figure~\ref{fig:coremke_correct} which illustrates an example where other methods make errors, while \method{} correctly answers the question.

\paragraph{Performance across n-hops:} We compare the performance of \mello{}, \pokemqa{} and \method{} in answering n-hop questions, $n \in {2, 3, 4}$ using \dataset{}-CF dataset and ChatGPT as the LLM. As shown in Fig.~\ref{fig:hop_ablation_avg}, \method{} outperforms \pokemqa{}-CL and MeLLo-CL with an average performance increase of 30.7\% for 2-hop questions, 22.6\% for 3-hop questions, and 5\% for 4-hop questions. Fig.~\ref{fig:hop_ablation_langs} presents language-wise accuracies for these methods for n-hop questions, showing the superior performance of \method{} compared to other methods.

\begin{figure}[h] 
  \centering
  \includegraphics[width=0.9\columnwidth]{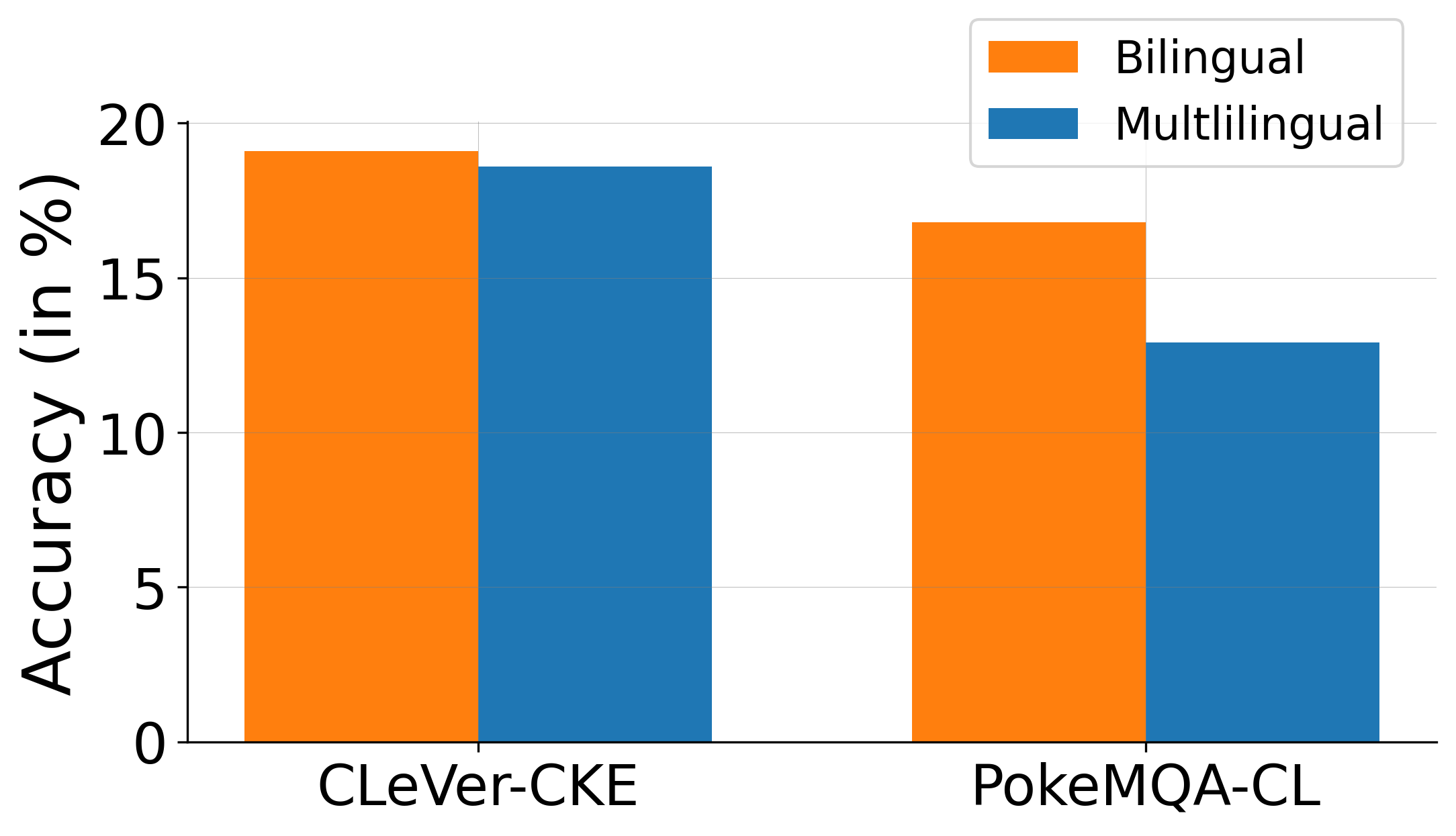} 
  \caption{Average accuracy using bilingual vs multilingual retriever, on the \dataset{}-CF dataset in 3k setting using ChatGPT as the LLM.}
  \label{fig:biling_multiling_pokemqa_ours}
  \vspace{-0.3cm}
\end{figure}

\paragraph{Bilingual vs Multilingual retriever:}
To compare performance differences with increasing the number of languages, we trained \pokemqacl{} and \method{}'s retrievers in a bilingual setting using English and the target language. See Fig \ref{fig:biling_multiling_pokemqa_ours} for results. As expected, on average the bilingual setting has greater performance than the multilingual setting, potentially due to interference of multiple languages in the multilingual setting. We interestingly observe that this gap is minimal in the case of \method{}, compared to \pokemqacl{}. This is because \method{}'s losses lead to better cross-lingual knowledge transfer leading to reduced interference of languages and more generalization. This observation generalizes across LLMs and datasets we tested on. Language-wise performance comparison of the two retriever setups for \pokemqa{} and \method{} using ChatGPT, \llama{}-7B and \vicuna{}-7B are in Tables~\ref{tab:chatgpt-cf}-\ref{tab:vicuna-t}. Also see Figs. \ref{fig:llama_acc_pokemqa} to \ref{fig:chatgpt_hopacc_ours} for more results.
\paragraph{Ablations:} We conducted an ablation on the loss functions we use, with results presented in Table~\ref{t:ablation_loss}. We selected five languages for this study and used the validation set of \dataset{}-CF. $\mathcal{L}_{SD}$ and $\mathcal{L}_{CLEC}$ significantly improve performance over $\mathcal{L}_{BCE}$, showing their importance in inducing fine-grained understanding and cross-lingual awareness in the retriever. Combining both all three losses leads to a 75.3\% and 109.5\% increase in average accuracy and hop-accuracy.

\begin{table}[ht]
    \centering
    \small
    \begin{tabular}{lccccc}
        \toprule
        Loss $\downarrow$ Lang. $\rightarrow$ & EN & DE & HI & SW & RU \\
        \midrule
        $\mathcal{L}_{BCE}$ &  26.0 & 28.0 & 16.0 & 20.0 & 16.0 \\
        + $\mathcal{L}_{SD}$ & 44.0 & 34.0 & 12.0 & 38.0 & 16.0 \\
        + $\mathcal{L}_{CLEC}$ & 44.0 & 36.0 & \textbf{18.0} & 30.0 & 18.0 \\
        + $\mathcal{L}_{SD}$ + $\mathcal{L}_{CLEC}$ & \textbf{76.0} & \textbf{62.0} & 12.0 & \textbf{58.0} & \textbf{26.0} \\
        \bottomrule
    \end{tabular}
    \caption{Ablation results of different loss functions used to train the retriever. Results on the validation set from \textit{\dataset-CF}.}
    \label{t:ablation_loss}
\end{table}

\noindent\textbf{Error analysis} We performed an error analysis of our method similar to the error analysis conducted for \pokemqa{}-CL and Mello-CL. We analyzed 30 samples each for edits made in English and Hindi. For English, based on random subset, we found that 70\% of the samples were correct, 8.1\% had \underline{Incorrect Retrieval} error, and 21.9\% had \underline{Incorrect LLM Response} error. In the case of Hindi, 46.6\% of the samples were correct. Of the remaining samples, 26.6\% had  \underline{Incorrect Retrieval} error, 16\% had both \underline{Incorrect LLM Response} and \underline{Incorrect Retrieval} errors, and 10.6\% had an \underline{Incorrect LLM Response} error. Refer Section~\ref{app:error} for more details. 
\section{Related Works}

\noindent\textbf{Knowledge editing methods:}  Knowledge editing can be broadly classified intro two groups. 1) Parameter-modifying based editing which locates the parameters related to factual knowledge and subsequently modify them~\citep{de-cao-etal-2021-editing, knowledgeneurons, mend, ROME, memit}.  These method requires an error-prone analytic step to identify parameters, which might be model-specific and not efficient. 2) Parameter-preserving based editing keeps the model parameters frozen and explicitly stores the fact edits in an external memory, for retrieval and external validation~\citep{zhong2023mquake, gu2024pokemqa, serac, hartvigsen2023aging}. Some recent works like that of \citet{hernandez2023measuring} have also explored a decoding time approach for editing knowledge. Further, knowledge editing is also explored in multimodal settings, such as for text-to-image models \citep{basu2023localizingeditingknowledgetexttoimage, basu2024mechanisticknowledgelocalizationtexttoimage, xiong2024editingmassiveconceptstexttoimage, gu2024pioneeringreliableassessmenttexttoimage}.

\noindent{\textbf{Cross-lingual knowledge editing.}} Recent studies have shifted focus to the multilingual capabilities of SoTA LLMs like LLaMA \cite{llama}, ChatGPT \cite{chatgpt}, and GPT-4 \cite{gpt4}. \citet{wang2023crosslingual} investigated cross-lingual knowledge editing and its impact on different target languages using a synthetic dataset. \cite{si2024mpn} introduced Multilingual Patch Neuron (MPN) for efficient cross-lingual knowledge synchronization, showing enhanced performance on single-hop XNLI and XFEVER datasets. \cite{xu2023language} proposed a framework for language anisotropic editing, facilitating simultaneous cross-lingual model editing. \cite{beniwal2024crosslingual} explored the cross-lingual model editing (XME) paradigm, revealing performance limitations in multilingual LLMs for hypernetwrok based parameter-modifying methods. \cite{wang2023retrievalaugmented} presented Retrieval-augmented Multilingual Knowledge Editing (ReMaKE), a model-agnostic knowledge editing method designed for multilingual settings. ReMaKE retrieves new knowledge from a multilingual knowledge base and concatenates it with prompts to update LLMs. Most of the above works have considered a simplistic setting of assuming the edited facts as independent without any multi-hop consequences of these edits, and are primarily focused on parameter updating based methods. We focus on parameter-preserving methods, and the more complex setting of multi-hop editing in a cross-lingual setup.

\noindent\textbf{Multi-Hop QA and prompting methods:} With the advances in generative language technologies powered by Large Language Models~\citep[LLMs;][]{gpt3, rae2022scaling, chowdhery2022palm, openai2023gpt4, tay2023ul2, Anil2023GeminiAF}, complex and multi-hop QA tasks are often handled by a prompt based and retrieval augmented approach \cite{press2022measuring, yao2022react, khattab2022demonstrate}. Works that tackle multi-hope knowledge editing have started to use this retrieve-then-generate framework to effeciently peform knowledge editing in an in-context setting, without changing the parameters of the base LLM, and have achieved SoTA performance on knowledge editing. Given their success, we use a similar retrieve, verify and generate strategy for knowledge editing with \method{}, while explicitly focussing on the retriever for enhanced knowledge editing performance.

\section{Conclusion}
In this paper, we contributed a benchmark having parallel multilingual examples for evaluating cross-lingual multi-hop knowledge editing. We provide new baselines and a detailed analysis of SoTA knowledge editing methods and find various gaps in existing methods, particularly in the cross-lingual setting. Motivated by this, we propose a generic, simple and highly effective method, \method{}, for improving the knowledge editing capabilities of parameter-preserving, retrieval augmented editing methods. \method{} improves cross-lingual and fine-grained retrieval in the case of knowledge editing, by introducing language aware and hard-negative mining based contrastive losses to train retrievers. Improved retrieval leads to precise knowledge retrieval and reduced error propagation in the multi-hop reasoning setting. \method{} is parameter-preserving in terms of the LLM weights, and uses a lightweight retriever with low latency as compared to methods like \citet{zhong2023mquake}.
\section{Limitations}
Our analysis and methods has some limitations. Firstly, although \dataset{} is a parallel cross-lingual benchmark, it predominantly contains fact edits related to English-speaking knowledge changes, while the edits could be localized to any part of the world in practice. This reliance on translation rather than trans-localization may lead to gaps in accurately understanding regional and local fact edits. However, having parallel data in all languages is advantageous to accurately measure per-language performance without confounding factors.
Secondly, our method is primarily focused on the retriever component and does not address the inherent inaccuracies of the LLMs. This includes issues such as understanding and generation capabilities of LLMs in different languages, correctly breaking down multi-hop questions into sub-questions, accurately extracting the final answer in the desired language.
Lastly, our analysis is currently limited to a broad range of medium to high-resource languages. Extending this analysis to low-resource languages presents a significant challenge due to the inaccuracies in translation, which can hinder the proper representation and understanding of facts in low resource languages.  Improving translation accuracy and extending our work to low-resource languages is part of our future work.
\section{Acknowledgements}
This work is, in part, supported by NSF (\#CNS-2431516). We thank the anonymous reviewers for their constructive feedback on this work. The
views, opinions, and/or findings contained in this paper are those of the authors and should not be interpreted as representing any funding agencies.

\bibliography{anthology,custom}
\bibliographystyle{acl_natbib}

\appendix

\section{Appendix}
\label{sec:appendix}

\subsection{Verification of Translated Data in \dataset{}}
\label{app:translation_verif}
\subsubsection{Human Verification of Translation}
\label{app_human_rating_translation}
We randomly selected 50 edits in four languages—German, Chinese, Hindi, and Bengali—and had the translations verified by expert human annotators to ensure accuracy. For each sample, we provided two sentences: one in English and its translation in the respective language. The annotators were asked to verify whether the semantic information was consistent between the two sentences. Given the brevity of the edit sentences, the potential for translation errors was minimal. Only one sample from Hindi in the \dataset{}-CF dataset encountered an issue during translation due to a special character error; the remaining samples were successfully processed. The expert human annotators suggested only minor stylistic changes for 1-2 words out of all 50 edit sentences in one language.

\subsubsection{Verification of Translations via Backtranslation}
\label{app:backtranslation}
To ensure the quality of translations, we employed back-translation, converting the translations from other languages back into English, and then calculated the average BLEU scores for 50 samples with the original English sentence as the ground truth. Table~\ref{tab:bleu_scores} presents these BLEU scores, indicating that six out of seven languages exhibit translations of very high quality, adequacy, and fluency \footnote{\url{https://cloud.google.com/translate/automl/docs/evaluate\#interpretation}}. For Chinese, the BLEU score suggests that the gist is clear, although there are some grammatical errors. However, with the addition of human verification (an expert gave a 100\% score to the translations in terms of preserving semantic content), we can conclude that the semantic information is preserved in the data translated to Chinese.

\begin{table}[h]
\centering
\begin{tabular}{lc}
\toprule
\textbf{Language} & \textbf{BLEU Score} \\
\midrule
\textbf{de} & 70.6 \\
\textbf{hi} & 59.2 \\
\textbf{bn} & 49.7 \\
\textbf{es} & 71.7 \\
\textbf{sw} & 65.9 \\
\textbf{ru} & 40.0 \\
\textbf{zh} & 23.0 \\
\bottomrule
\end{tabular}
\caption{BLEU Scores for back-translation to English for different languages.}
\label{tab:bleu_scores}
\end{table}

\subsection{Training Details}
We employ the training dataset to train the retriever component of the \method{} framework, using the same training set as utilized in training \pokemqacl{} \cite{gu2024pokemqa}. Subsequently, we translate this dataset into seven other languages and generate hard negatives following the method outlined in Section~\ref{method}. The training dataset contains 6688 samples along with translations into 8 langugaes and hard-negative pairs for each edit in the dataset, both of which is created by us for training \method{}'s retriever. For training the multilingual retriever, we utilize data from all languages, while for training the bilingual retriever, we focus on English and the target language data. To optimize our method's performance, we conduct hyperparameter tuning on a validation set derived from \dataset{}-CF, comprising 50 samples exclusively for this purpose without involvement in inferencing tasks. The hyperparameters used for tuning are mentioned in Table~\ref{tab:hyperparameters}. Our experiments are expensive (See Appendix ~\ref{app:resources}) and we do not perform experiments on multiple seeds.

\subsection{Method Details}
We finetuned distilbert-base-multilingual-cased \cite{Sanh2019DistilBERTAD} with approximately 130.7M parameters from the HuggingFace transformers library on the training data we created by translation and hard negative mining for the edits as described in Section~\ref{method} using our designed training objectives for the retriever. We used held out 20\% of the samples for the validation set and used Adam optimizer to update the parameters during training. 
\begin{table}[h]
\centering
\begin{tabular}{ll}
\toprule
\textbf{Hyperparameter} & \textbf{Value} \\
\midrule
Learning Rate  & $5.00 \times 10^{-5}$ \\
Batch Size     & \{1024, 2048\} \\
Epoch & 200 \\
Margin         & 1 \\
Threshold & \{0.5 , 0.7\} \\
\bottomrule
\end{tabular}
\caption{Hyperparameter values searched for tuning the multilingual retriever in and \method{} and \pokemqacl{}.}
\label{tab:hyperparameters}
\end{table}

\subsection{\dataset{} Benchmark Statistics}
See Table \ref{t:data_stats} for the dataset statistics of our benchmark \dataset{}, which we create in this work and use it for evaluating the cross-lingual multi-hop knowledge editing capabilities of various model editing techniques.
Languages studied in this work and supported by \dataset{} are English, German, Spanish, Hindi, Swahili, Bengali, Russian, Chinese.

\begin{table*}[h!]
    \centering
    \resizebox{0.95\textwidth}{!}{
    \begin{tabular}{ccccccc}
        \toprule
        & \#Edits & \multicolumn{4}{c}{Hop-Wise Stats (per-language/total)} & \#Languages\\
        \cmidrule(lr){3-6}
        & & 2-hop & 3-hop & 4-hop & Total & \\
        \midrule
        \multirow{5}{*}{\dataset{}-CF} & 1 & 513 / 4k & 356 / 2.8k & 224 / 1.8k & 1093 / 8.7k & 8 \\
        & 2 & 487 / 3.9k & 334 / 2.7k & 246 / 2k & 1067 / 8.5k & 8 \\
        & 3 & - & 310 / 2.5k & 262 / 2.1k & 572 / 4.6k & 8 \\
        & 4 & - & - & 268 / 2.1k & 268 / 2.1k & 8 \\
        & All & 1000 / 8k & 1000 / 8k & 1000 / 8k & 3000 / 24k & 8 \\
        \midrule
        \dataset{}-T& 1 (All) & 1421 / 11368 & 445 / 3560 & 2 / 16 & 1868 / 14944 & 8\\
        \bottomrule
    \end{tabular}
    }
    \caption{Statistics of \dataset{} created and used in our experiments. Statistics per language are same as those reported in \citet{zhong2023mquake}.}
    \label{t:data_stats}
\end{table*}

\subsection{Prompts for LLM inference}
\label{prompts}
To help the LLM break down questions into subquestions, generate answers for the subquestions, and extract the final answer, we provide four in-context example demonstrations. These examples include edits from different languages based on the edits made. We include a mix of 2, 3, and 4-hop example demonstrations in the prompt.  Below, we present an example demonstration for a prompt used for edits in German and Swahili. In these demonstrations, the text written in \textcolor{blue}{blue} represents the updated fact from the edited fact memory, and the text written in \textcolor{teal}{teal} indicates the answer extraction.  

Here is the 3-hop question example demonstration used in the prompt when edits are made in German: \\

\noindent\textit{Question: What is the capital city of the country of citizenship of Ivanka Trump's spouse? \\
Subquestion: Who is Ivanka Trump's spouse? \\
Generated answer: \textcolor{blue}{Der Ehemann von Ivanka Trump ist Jared Kushner.}\\
According to Generated answer, the entity of Subquestion in English is: \textcolor{teal}{Jared Kushner}\\
Subquestion: What is the country of citizenship of Jared Kushner?\\
Generated answer: \textcolor{blue}{Jared Kushner ist kanadischer Staatsbürger.}\\
According to Generated answer, the entity of Subquestion in English is: \textcolor{teal}{Canada}\\
Subquestion: What is the capital city of Canada?\\
Generated answer: \textcolor{blue}{Die Hauptstadt Kanadas ist Ottawa.}\\
According to Generated answer, the entity of Subquestion in English is: \textcolor{teal}{Ottawa}.\\
Final answer: \textcolor{teal}{Ottawa} \\
}

Following is the 2-Hop example demonstration when edits are made in Swahili: \\

\noindent\textit{
Question: Who is the head of state of the country where Rainn Wilson holds a citizenship? \\
Subquestion: What is the country of citizenship of Rainn Wilson? \\
Generated answer: \textcolor{blue}{Rainn Wilson ni raia wa Kroatia.} \\
According to Generated answer, the entity of Subquestion in English is: \textcolor{teal}{Croatia} \\
Subquestion: What is the name of the current head of state in Croatia? \\
Generated answer: \textcolor{blue}{Jina la mkuu wa sasa wa nchi nchini Kroatia ni Kolinda Grabar-Kitarović.} \\
According to Generated answer, the entity of Subquestion in English is: \textcolor{teal}{Kolinda Grabar-Kitarović} \\
Final answer: \textcolor{teal}{Kolinda Grabar-Kitarović} \\
}

\subsection{Compute Resources}
\label{app:resources}
We performed all experiments using 8 NVIDIA A100 80 GB GPUs. The training duration for the retriever, including both bilingual and multilingual retrievers for both \pokemqacl{} and \method{}, was approximately 2 hours per run. Inference tasks took between 4 to 6 hours to complete when using ChatGPT as the LLM in the case of \method{}, and between 10 to 24 hours with Llama-2-7b and Vicuna-1.5. Each MeLLo baseline run varied in duration from 8 to 24 hours, depending on the language and the LLM used.

\subsection{Error Analysis}
\label{app:error}
\begin{figure*}
    \centering
  \includegraphics[width=\textwidth]{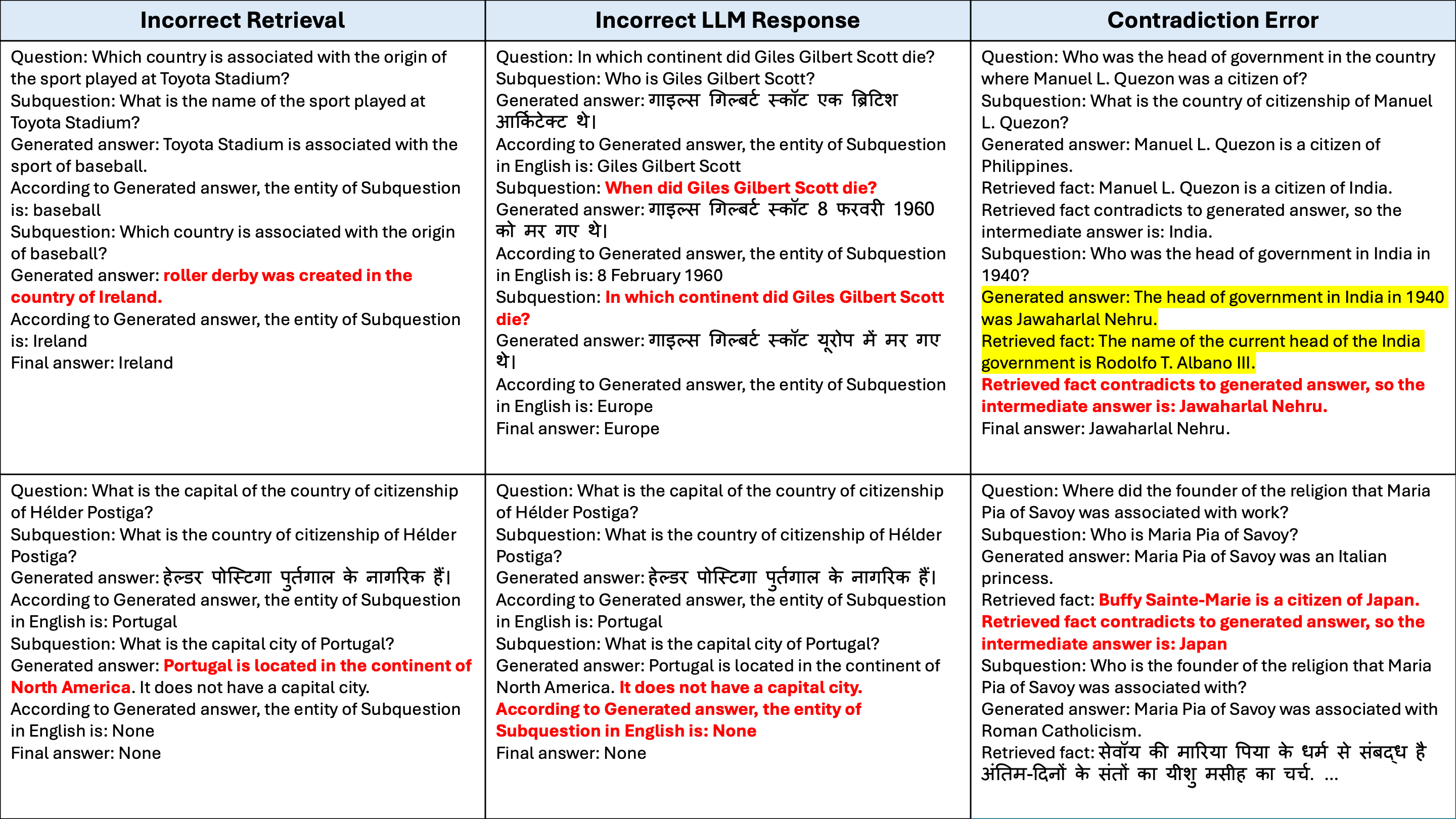} 
  \caption{Examples of types of errors made by different methods such as \mellocl{}, \pokemqacl{} and \method{}. Text in red highlights the step at which the error is made. Text highlighted in yellow means the steps that are correct but lead to error in contradiction. Examples are provided in English and Hindi.}
  \label{fig:error_types}
\end{figure*}

Figure~\ref{fig:error_types} presents real examples of errors made by different methods. The first column displays errors related to incorrect retrieval, where the model fails to understand the context of the subquestion and either retrieves a fact with some word overlap with the subquestion or a random edit. The second column shows instances where the LLM makes mistakes in breaking down the subquestion. In the first example, it deviates from the question, asking \textbf{when} Giles Gilbert Scott died, and then in the third hop, it just repeats the original question. The second example of this column contains an example where the LLM fails to adhere to the strict pattern of the prompt, misunderstands the context, and generates incorrect information, causing a cascading effect of errors. The third column highlights errors specific to the \mello method, where the LLM struggles to disambiguate between the generated answer and the retrieved fact. In the first example of this column, the retrieved fact contradicts the generated answer, but the LLM fails to identify the correct entity from the generated answer/retrieved fact after resolving the contradiction, leading to a wrong answer. In the second example, although the retrieved fact does not contradict the generated answer, the LLM incorrectly perceives it as a contradiction, resulting in a mistake.

\begin{figure*}
    \centering
  \includegraphics[trim=0 110 0 0, clip, width=\textwidth]{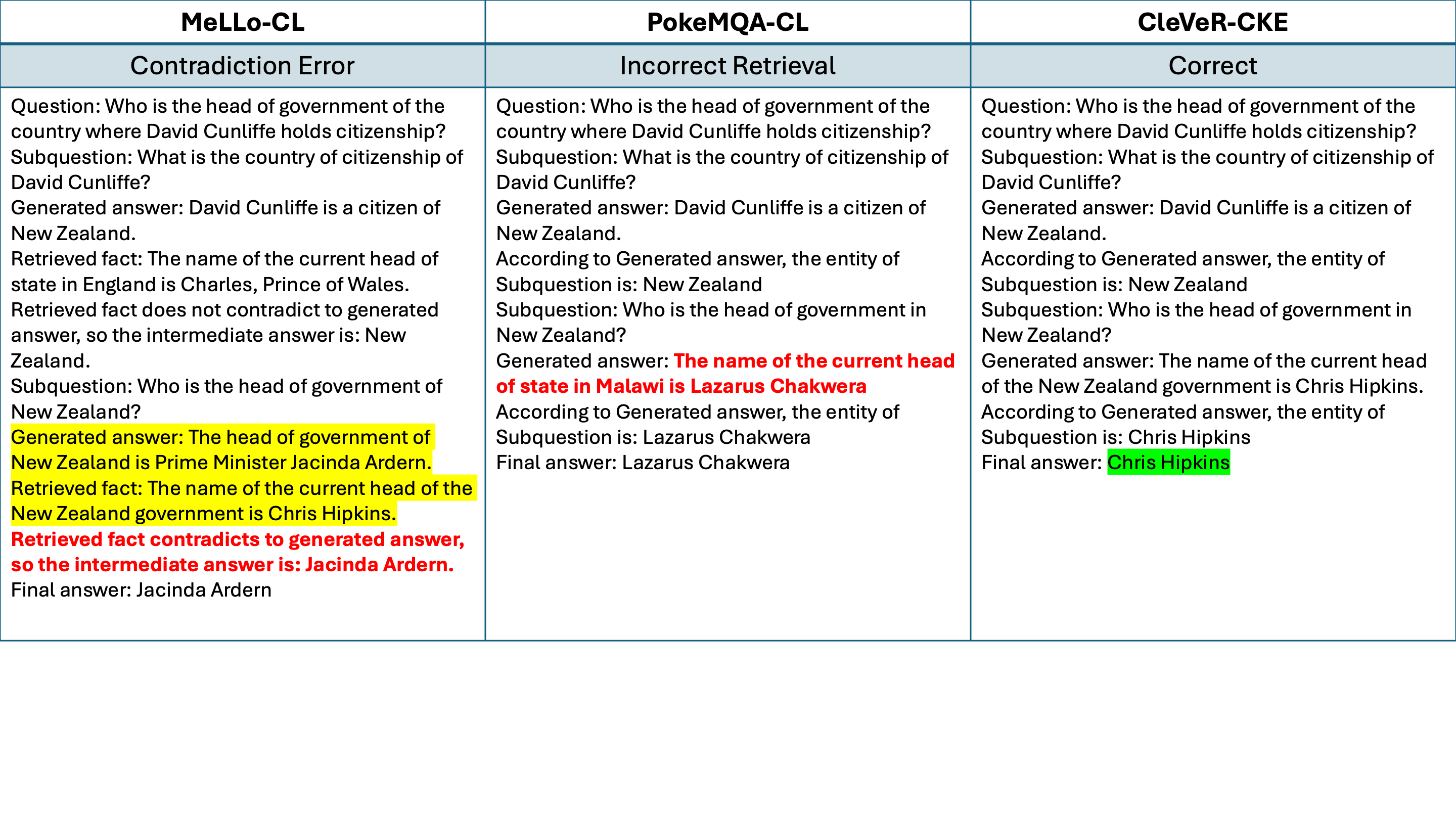} 
  \caption{Sample of data showing how \method{} doesn't make the errors of MeLLo-CL and \pokemqacl{}-CL. Text in red highlights the step at which the error is made. Text highlighted in yellow means the steps that are correct but lead to error in contradiction. Text highlighted in green means the correct final answer achieved by taking all correct steps.}
  \label{fig:coremke_correct}
\end{figure*}

Our method, \method{}, addresses and improves upon these errors, as demonstrated in Figure~\ref{fig:coremke_correct}. In the same question scenario, where \mellocl{} exhibits a contradiction error highlighted in yellow and red, and \pokemqacl{} makes a retrieval error based on word overlap, our method follows all the correct steps, leading to the accurate final answer.

\subsection{Licensing}

The baseline methods ROME, MEMIT, FT, \mello, and \pokemqa{} are distributed under the MIT License. Similarly, the datasets MQUAKE-CF and MQUAKE-T are available under the MIT License. The models \vicuna{}-7B (v1.5) and distilbert-base-multilingual-cased are released under the Apache License 2.0, while \llama{}-7B is licensed under the LLAMA 2 Community License.

\begin{figure*}[h] 
    \includegraphics[width=\textwidth]{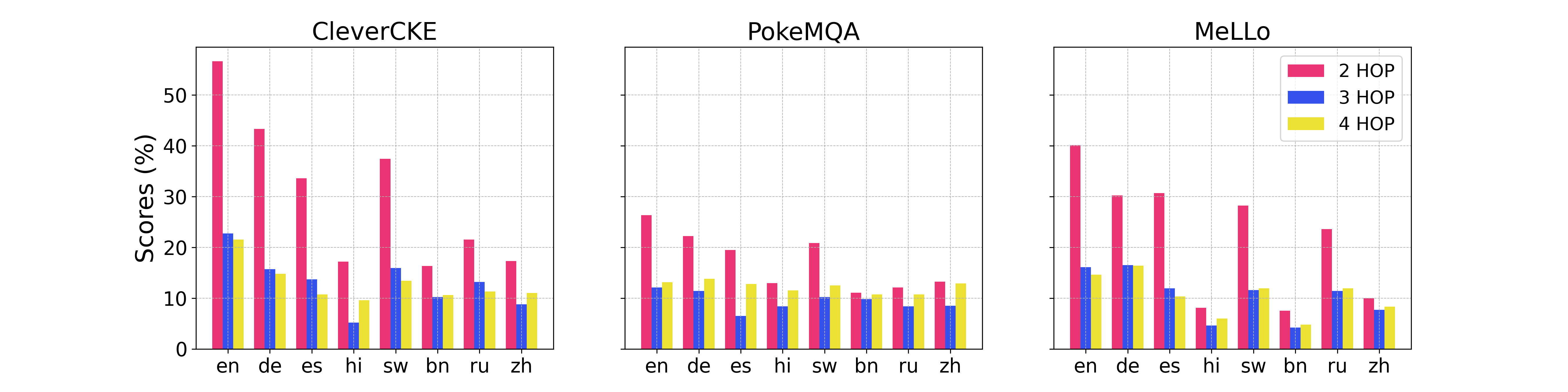}
  \caption{Accuracy of methods \method{}, \pokemqa{}-CL and MeLLo-CL reported on 2, 3, 4-hop questions in \dataset{}-CF with ChatGPT as LLM for all languages. We take the 3k edit case using \dataset{}-CF.}
  \label{fig:hop_ablation_langs}
\end{figure*}

\begin{figure*}[h]
    \centering
    \begin{minipage}[t]{0.45\textwidth}
        \centering
        \includegraphics[width=\textwidth]{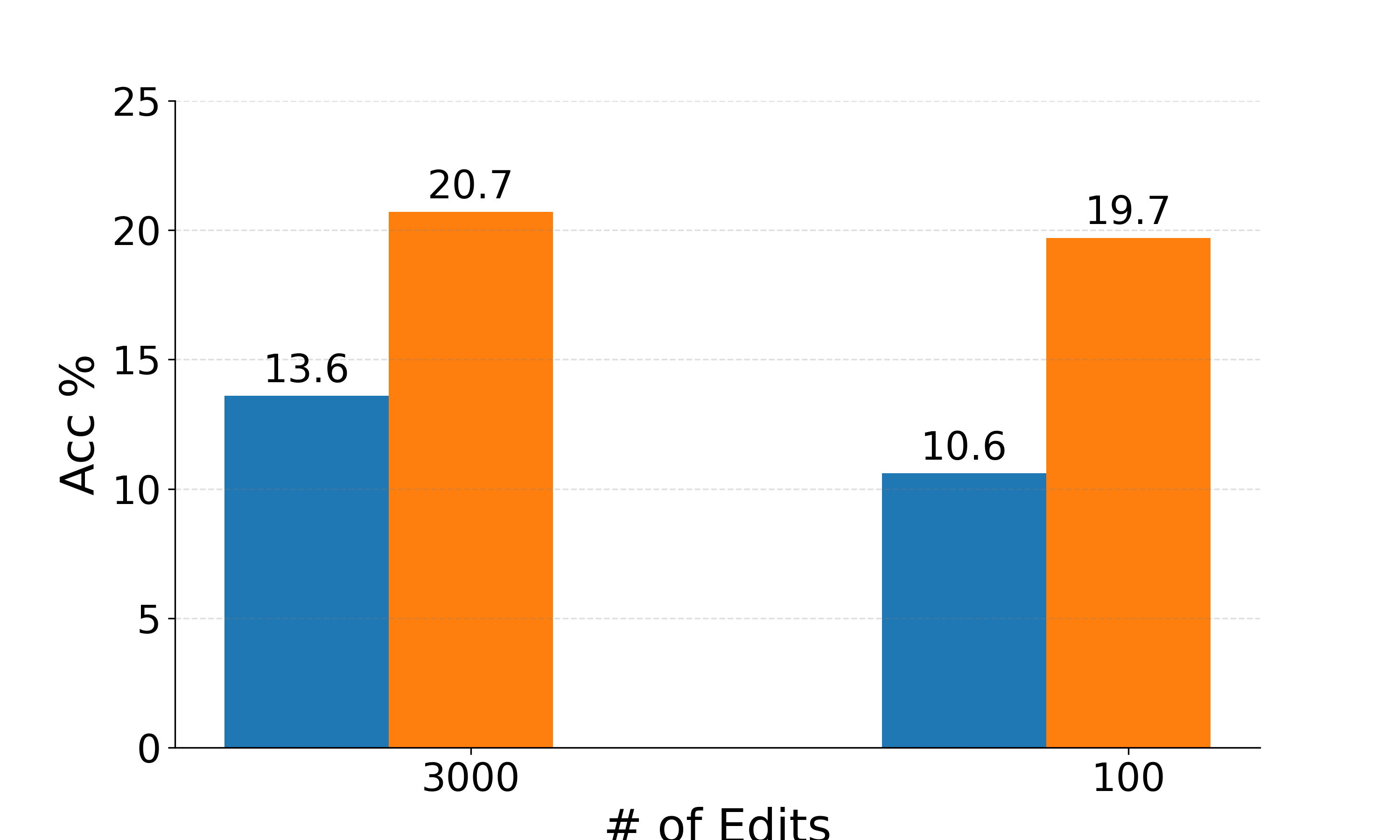}
        \caption{Knowledge Editing accuracy of \pokemqacl{} using \llama{} as the LLM in the Bilingual and Multilingual Case, for two cases -- edited fact memory size kept as 3k and 100 edits.}
        \label{fig:llama_acc_pokemqa}
    \end{minipage}\hfill
    \begin{minipage}[t]{0.45\textwidth}
        \centering
        \includegraphics[width=\textwidth]{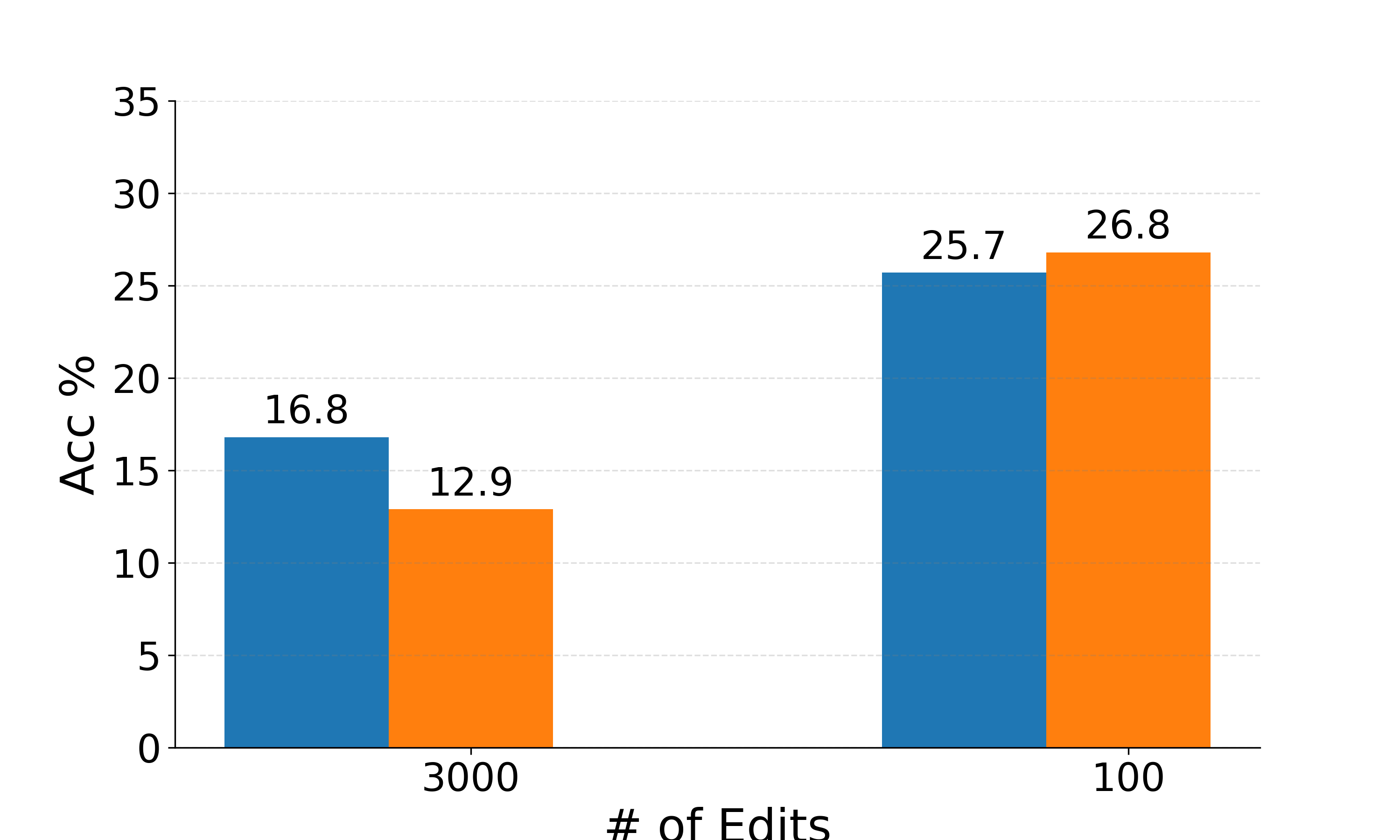}
        \caption{Knowledge Editing accuracy of \pokemqacl{} using \chatgpt{} as the LLM in the Bilingual and Multilingual Case, for two cases -- edited fact memory size kept as 3k and 100 edits.}
        \label{fig:chatgpt_acc_pokemqa}
    \end{minipage}
    
    \vspace{1em}
    
    \begin{minipage}[t]{0.45\textwidth}
        \centering
        \includegraphics[width=\textwidth]{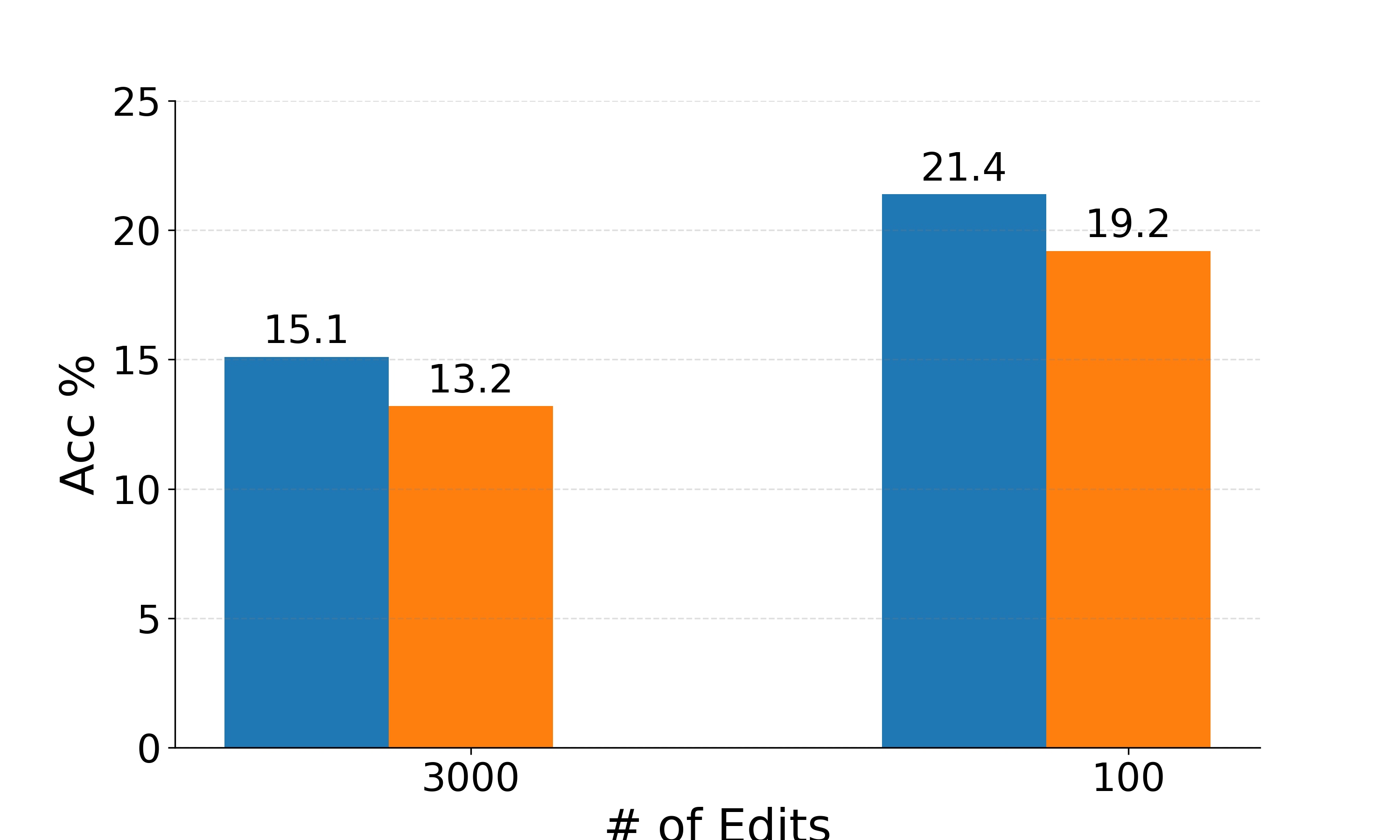}
        \caption{Knowledge Editing accuracy of \method{} using \llama{} as the LLM in the Bilingual and Multilingual Case, for two cases -- edited fact memory size kept as 3k and 100 edits.}
        \label{fig:llama_acc_ours}
    \end{minipage}\hfill
    \begin{minipage}[t]{0.45\textwidth}
        \centering
        \includegraphics[width=\textwidth]{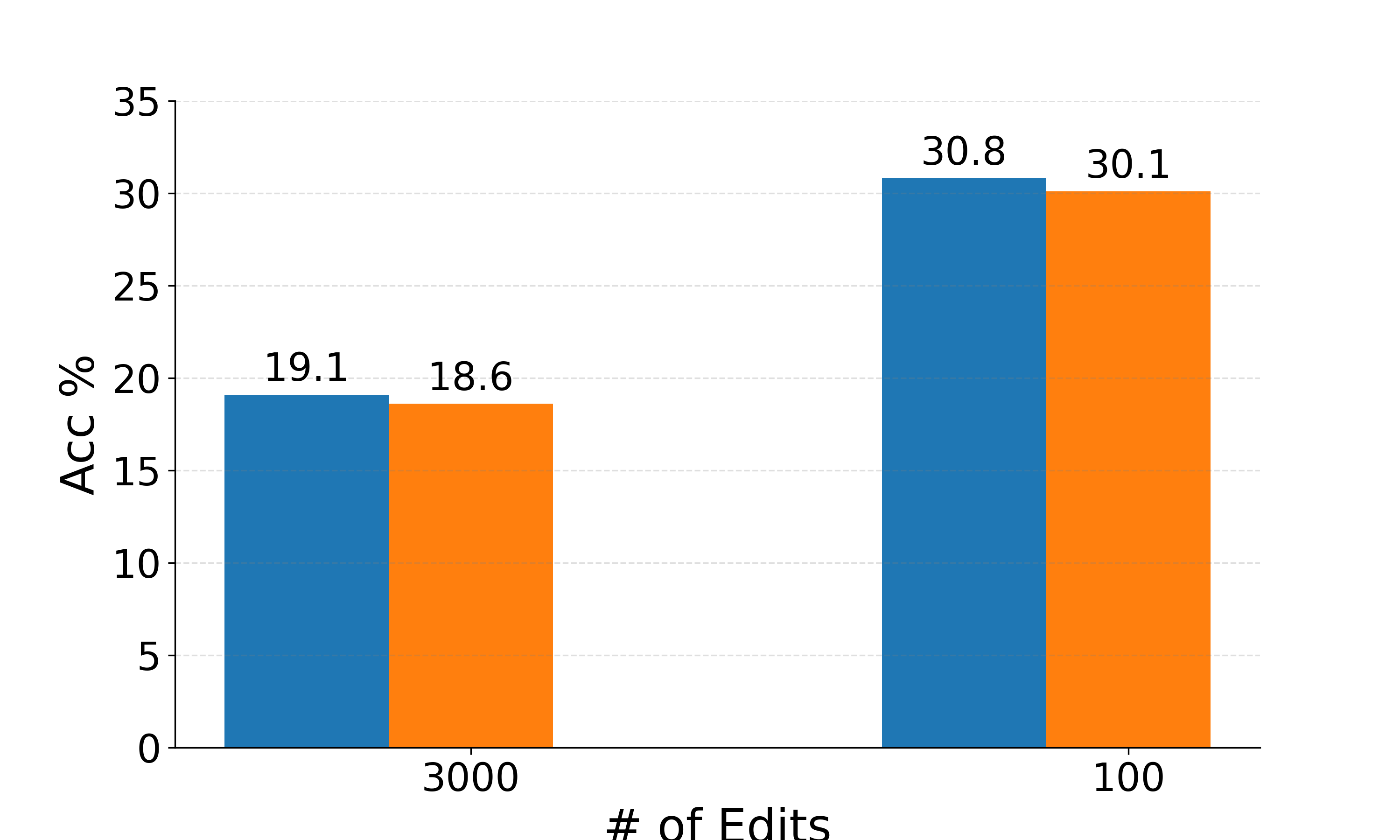}
        \caption{Knowledge Editing accuracy of \method{} using \chatgpt{} as the LLM in the Bilingual and Multilingual Case, for two cases -- edited fact memory size kept as 3k and 100 edits.}
        \label{fig:chatgpt_acc_ours}
    \end{minipage}
    
    \vspace{1em}
    
    \begin{minipage}[t]{0.45\textwidth}
        \centering
        \includegraphics[width=\textwidth]{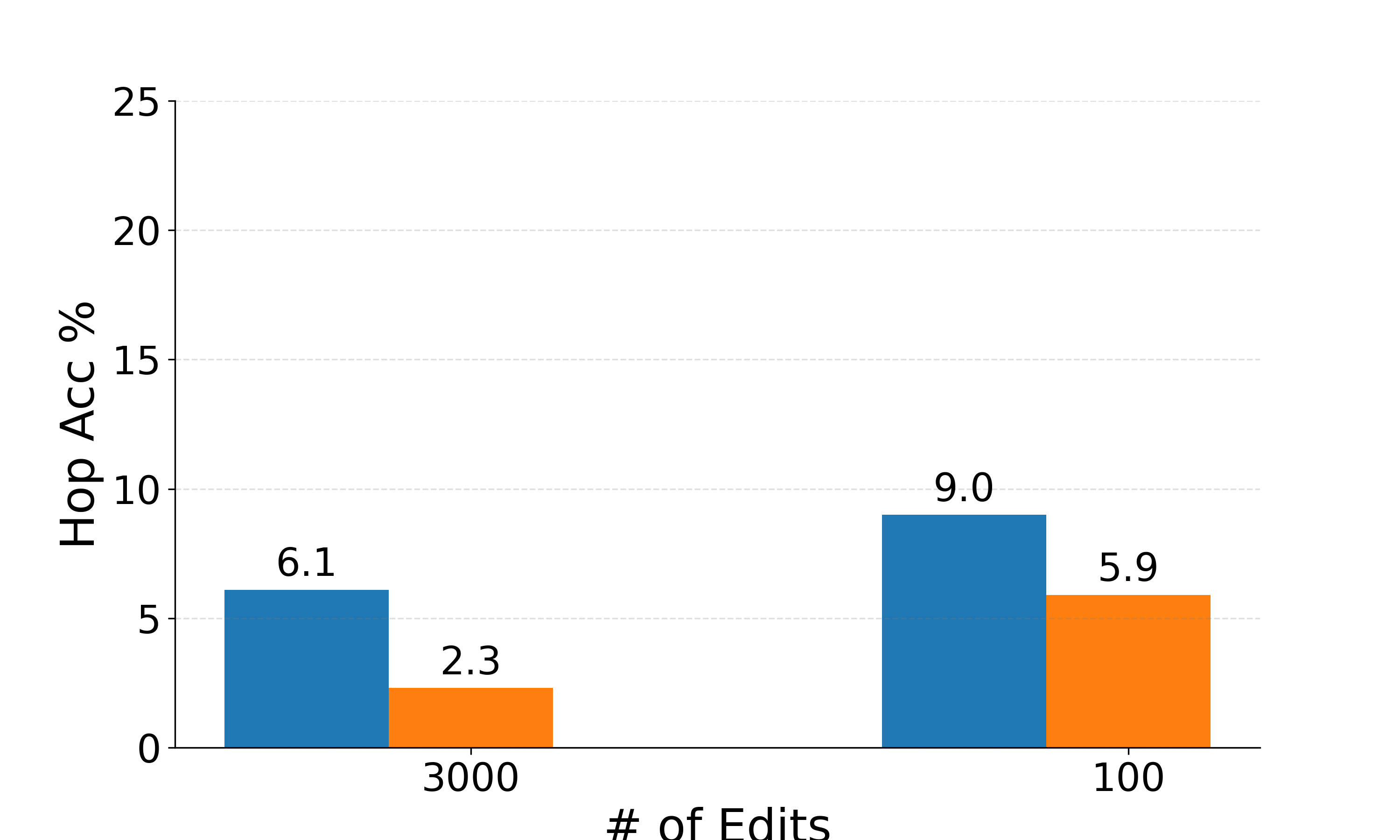}
        \caption{Hop-Accuracy of \pokemqacl{} using \llama{} as the LLM in the Bilingual and Multilingual Case, for two cases -- edited fact memory size kept as 3k and 100 edits.}
        \label{fig:llama_hopacc_pokemqa}
    \end{minipage}\hfill
    \begin{minipage}[t]{0.45\textwidth}
        \centering
        \includegraphics[width=\textwidth]{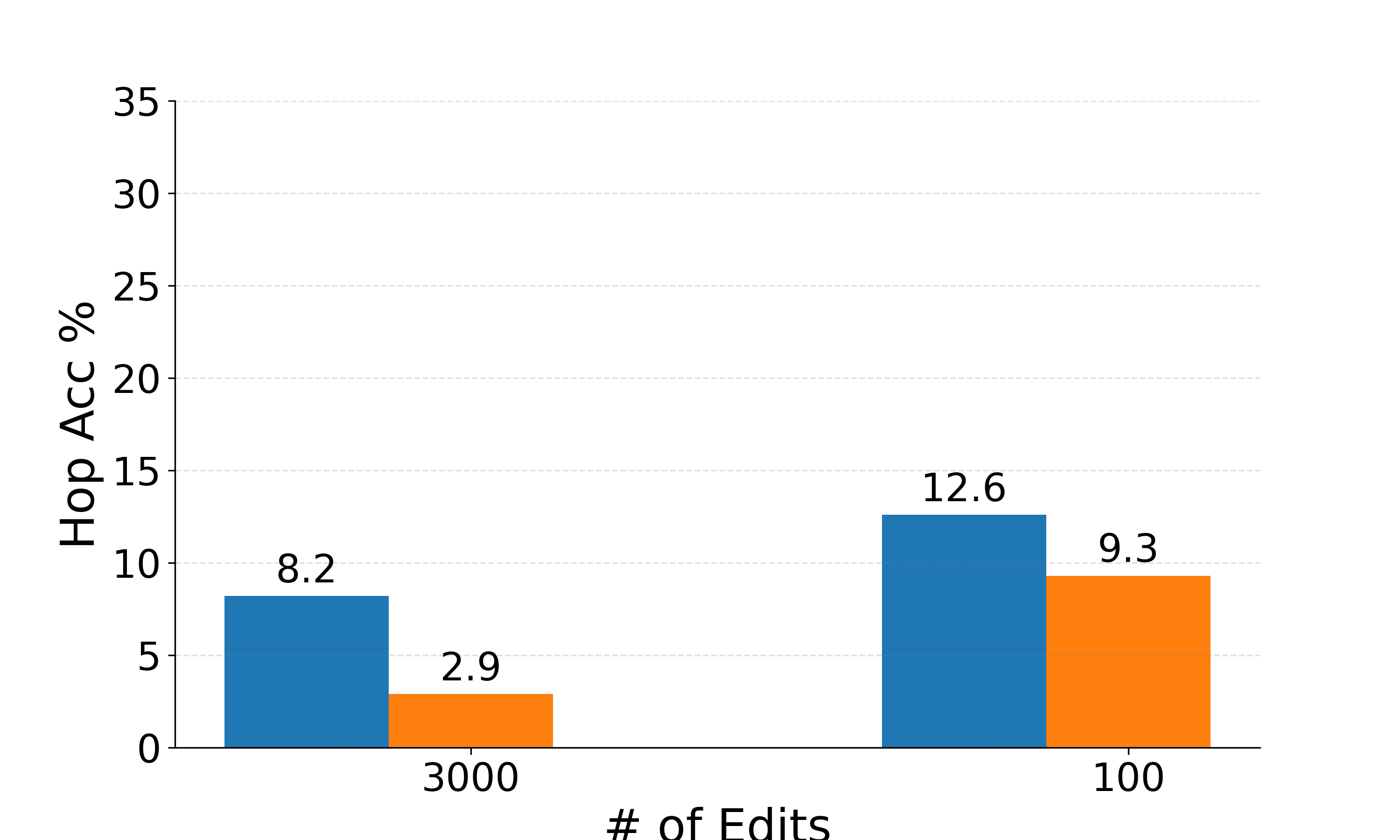}
        \caption{Hop-Accuracy of \pokemqacl{} using \chatgpt{} as the LLM in the Bilingual and Multilingual Case, for two cases -- edited fact memory size kept as 3k and 100 edits.}
        \label{fig:chatgpt_hopacc_pokemqa}
    \end{minipage}
    
    \vspace{1em}
    
    \begin{minipage}[t]{0.45\textwidth}
        \centering
        \includegraphics[width=\textwidth]{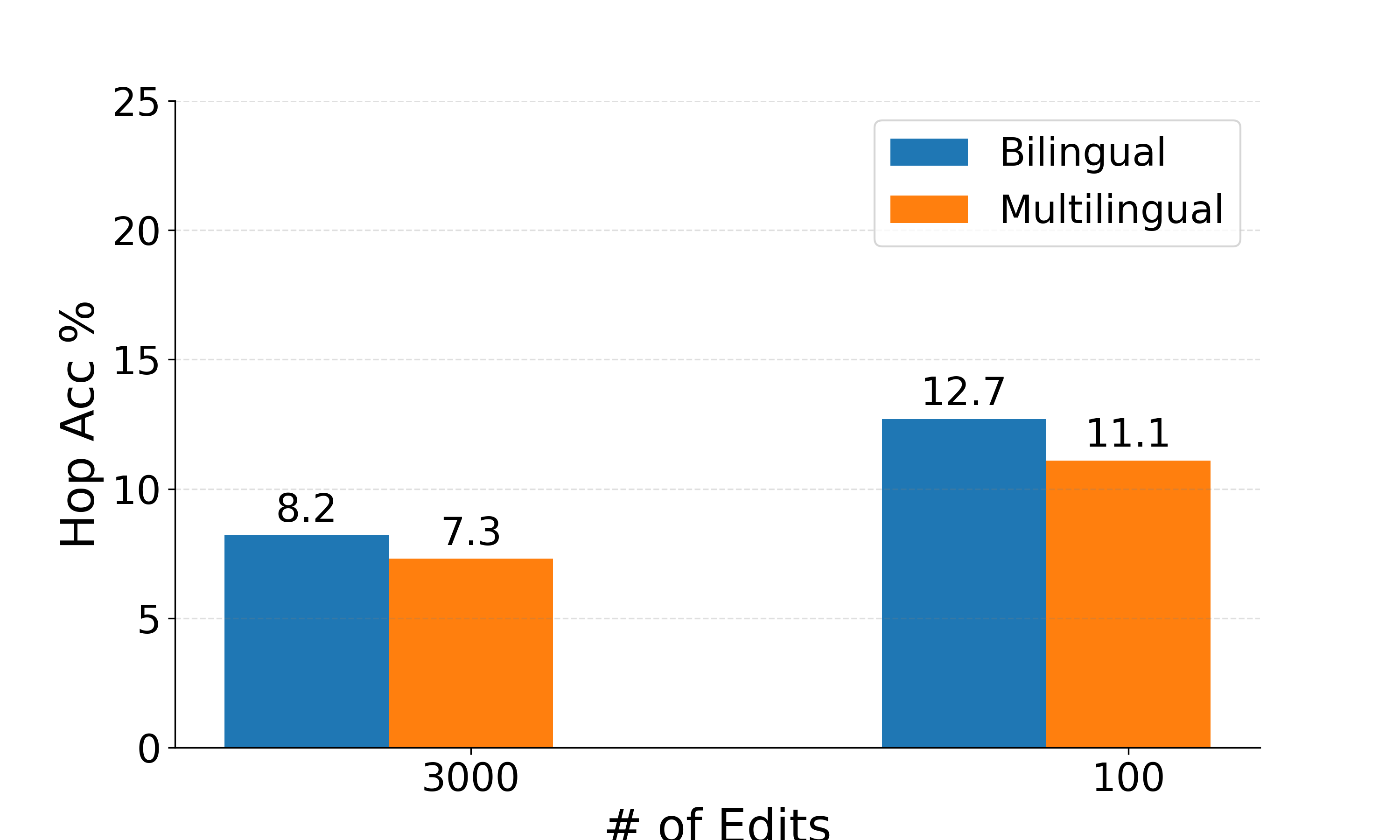}
        \caption{Hop-Accuracy of \method{} using \llama{} as the LLM in the Bilingual and Multilingual Case, for two cases -- edited fact memory size kept as 3k and 100 edits.}
        \label{fig:llama_hopacc_ours}
    \end{minipage}\hfill
    \begin{minipage}[t]{0.45\textwidth}
        \centering
        \includegraphics[width=\textwidth]{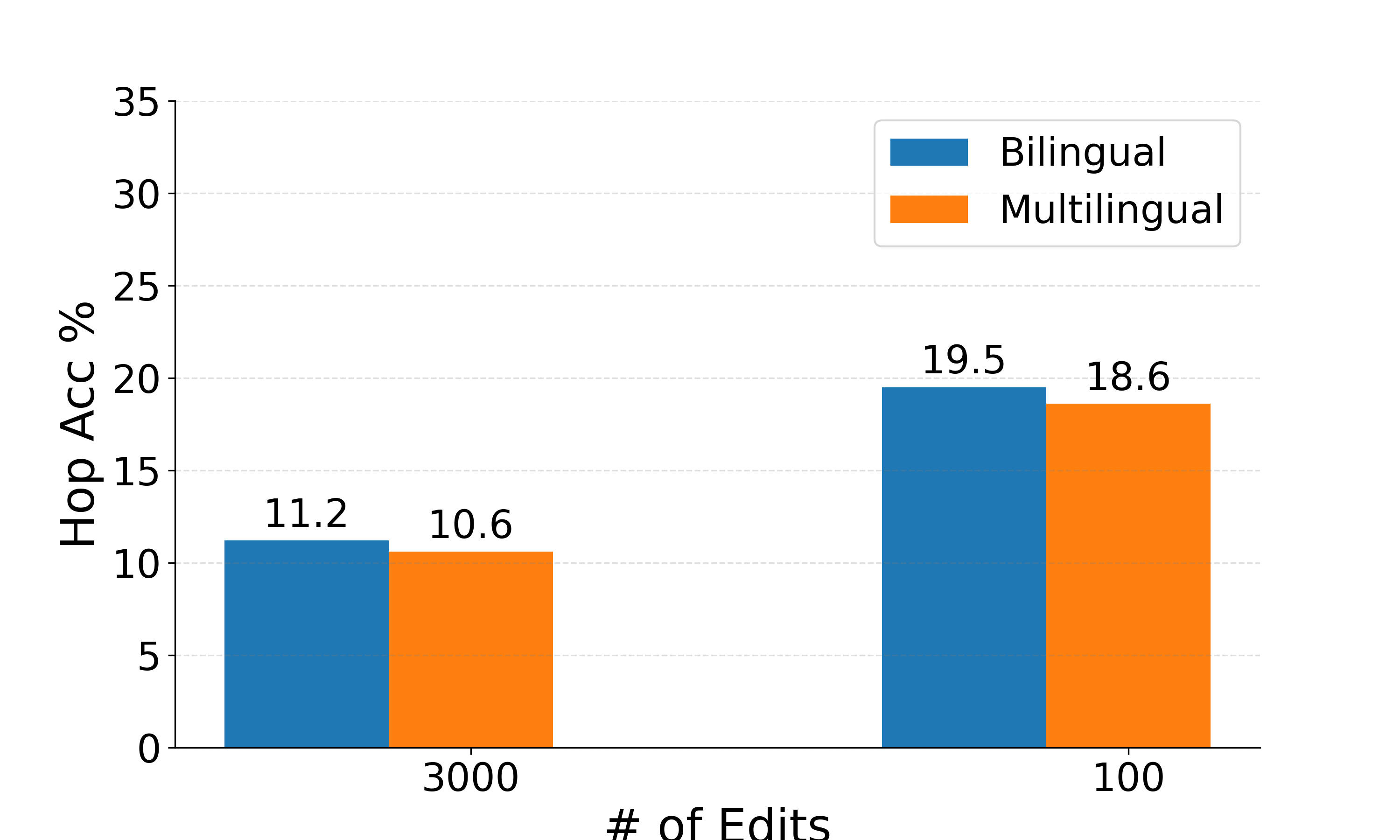}
        \caption{Hop-Accuracy of \method{} using \chatgpt{} as the LLM in the Bilingual and Multilingual Case, for two cases -- edited fact memory size kept as 3k and 100 edits.}
        \label{fig:chatgpt_hopacc_ours}
    \end{minipage}
    
\end{figure*}

\begin{table*}[h]
    \centering
    \begin{adjustbox}{max width=0.8\textwidth}
    \begin{tabular}{ccccccccccc}
        \toprule
        & \textbf{Edits} & \multicolumn{2}{c}{\textbf{Bilingual 3k}} & \multicolumn{2}{c}{\textbf{Multilingual 3k}} & \multicolumn{2}{c}{\textbf{Bilingual 100}} & \multicolumn{2}{c}{\textbf{Multilingual 100}} \\
        \cmidrule(lr){3-4} \cmidrule(lr){5-6} \cmidrule(lr){7-8} \cmidrule(lr){9-10}
        & & \textbf{Acc} & \textbf{Hop-Acc} & \textbf{Acc} & \textbf{Hop-Acc} & \textbf{Acc} & \textbf{Hop-Acc} & \textbf{Acc} & \textbf{Hop-Acc} \\
        \midrule
        \multirow{8}{*}{\begin{turn}{90}\pokemqacl{}\end{turn}} & en & 39.1 & 30.7 & 17.0 & 7.3 & 55.9 & 47.2 & 35.9 & 19.5 \\
& de & 25.1 & 14.5 & 15.7 & 3.7 & 29.3 & 16.6 & 33.0 & 12.5 \\
& es & 20.6 & 9.4 & 12.8 & 2.8 & 29.7 & 13.5 & 28.2 & 9.2 \\
& hi & 6.8 & 0.2 & 10.9 & 1.0 & 16.0 & 1.3 & 21.4 & 4.0 \\
& sw & 17.0 & 9.2 & 14.4 & 4.0 & 22.3 & 13.4 & 30.7 & 11.5 \\
& bn & 11.1 & 0.3 & 10.5 & 1.2 & 15.9 & 1.5 & 21.6 & 4.4 \\
& ru & 7.9 & 0.7 & 10.4 & 1.5 & 20.2 & 4.3 & 23.2 & 7.7 \\
& zh & 7.1 & 0.6 & 11.5 & 1.5 & 16.3 & 3.0 & 20.5 & 5.4 \\
        \midrule
        & \pokemqacl{} & 16.8 & 8.2 & 12.9 & 2.9 & 25.7 & 12.6 & 26.8 & 9.3 \\
        \midrule
        \multirow{8}{*}{\begin{turn}{90}\method{}\end{turn}} & en & 36.2 & 28.7 & 33.1 & 25.0 & 57.5 & 48.8 & 54.8 & 43.8 \\
&  de & 29.2 & 16.0 & 24.3 & 14.3 & 38.1 & 23.9 & 39.2 & 24.3 \\
&  es & 21.4 & 11.3 & 19.1 & 10.0 & 34.2 & 18.4 & 31.6 & 17.6 \\
&  hi & 10.5 & 4.9 & 10.5 & 4.4 & 22.8 & 10.6 & 17.3 & 8.2 \\
&  sw & 21.9 & 14.3 & 22.0 & 13.6 & 34.7 & 24.6 & 37.9 & 24.6 \\
&  bn & 12.0 & 4.5 & 12.3 & 4.3 & 16.8 & 7.8 & 16.8 & 7.1 \\
&  ru & 13.0 & 7.1 & 15.2 & 7.9 & 25.7 & 14.7 & 24.4 & 14.1 \\
&  zh & 8.6 & 3.1 & 12.3 & 5.4 & 16.5 & 6.8 & 19.2 & 9.5 \\
        \midrule
        & \method{} & \textbf{19.1} & \textbf{11.2} & \textbf{18.6} & \textbf{10.6} & \textbf{30.8} & \textbf{19.5} & \textbf{30.1} & \textbf{18.6} \\
        \bottomrule
    \end{tabular}
    \end{adjustbox}
    \caption{Performance of \pokemqacl{} and \method{} by Language and Number of Edits on the \dataset{}-CF Dataset Using ChatGPT Backbone: Bilingual and Multilingual Training of the Retriever with All and 100 Edits.}
    \label{tab:chatgpt-cf}
\end{table*}

\begin{table*}[h]
    \centering
    \begin{adjustbox}{max width=0.8\textwidth}
    \begin{tabular}{ccccccccccc}
        \toprule
        & \textbf{Edits} & \multicolumn{2}{c}{\textbf{Bilingual 1.8k}} & \multicolumn{2}{c}{\textbf{Multilingual 1.8k}} & \multicolumn{2}{c}{\textbf{Bilingual 100}} & \multicolumn{2}{c}{\textbf{Multilingual 100}} \\
        \cmidrule(lr){3-4} \cmidrule(lr){5-6} \cmidrule(lr){7-8} \cmidrule(lr){9-10}
        & & \textbf{Acc} & \textbf{Hop-Acc} & \textbf{Acc} & \textbf{Hop-Acc} & \textbf{Acc} & \textbf{Hop-Acc} & \textbf{Acc} & \textbf{Hop-Acc} \\
        \midrule
        \multirow{8}{*}{\begin{turn}{90}\pokemqacl{}\end{turn}} & en & 79.1 & 69.1 & 23.7 & 17.6 & 79.3 & 69.5 & 30.0 & 22.5 \\
& de & 45.1 & 32.3 & 13.7 & 08.9 & 46.5 & 33.5 & 17.7 & 11.1 \\
& es & 41.0 & 28.2 & 06.7 & 03.6 & 45.2 & 31.2 & 13.3 & 8.0 \\
& hi & 13.4 & 6.4 & 8.6 & 4.8 & 15.7 & 8.6 & 12.4 & 7.0 \\
& sw & 54.8 & 41.9 & 15.5 & 9.4 & 58.7 & 44.3 & 19.3 & 11.6 \\
& bn & 11.7 & 5.7 & 13.8 & 6.0 & 12.8 & 6.4 & 14.2 & 7.2 \\
& ru & 12.5 & 7.5 & 14.9 & 10.0 & 14.2 & 9.4 & 16.9 & 10.9 \\
& zh & 10.8 & 5.9 & 11.0 & 5.6 & 14.2 & 8.4 & 15.1 & 7.4 \\
        \midrule
        & \pokemqacl{} & 33.5 & 24.6 & 13.5 & 8.2 & 35.8 & 26.4 & 17.4 & 10.7 \\
        \midrule
        \multirow{8}{*}{\begin{turn}{90}\method{}\end{turn}} & en & 80.6 & 69.9 & 66.6 & 54.7 & 81.0 & 70.3 & 67.4 & 55.4 \\
        & de & 63.6 & 50.2 & 59.3 & 46.5 & 64.1 & 50.6 & 59.7 & 46.6 \\
        & es & 45.7 & 32.2 & 28.7 & 19.9 & 46.3 & 32.9 & 29.3 & 20.2 \\
        & hi & 39.3 & 25.6 & 17.0 & 9.6 & 42.0 & 27.2 & 16.8 & 9.5 \\
        & sw & 47.7 & 37.3 & 51.8 & 37.6 & 50.1 & 39.1 & 52.1 & 37.8 \\
        & bn & 20.7 & 14.1 & 14.3 & 8.3 & 20.9 & 14.2 & 14.5 & 8.5 \\
        & ru & 58.0 & 45.2 & 31.4 & 22.2 & 62.5 & 50.2 & 32.0 & 22.5 \\
        & zh & 46.6 & 34.3 & 35.7 & 23.3 & 49.0 & 35.7 & 35.6 & 23.2 \\
        \midrule
        & \method{} & \textbf{50.3} & \textbf{38.6} & \textbf{38.1} & \textbf{27.7}  & \textbf{52.0} & \textbf{40.0} & \textbf{38.4} & \textbf{28.0} \\
        \bottomrule
    \end{tabular}
    \end{adjustbox}
    \caption{Performance of \pokemqacl{} and \method{} by Language and Number of Edits on the \dataset{}-T Dataset Using ChatGPT Backbone: Bilingual and Multilingual Training of the Retriever with All and 100 Edits.}
    \label{tab:chatgpt-t}
\end{table*}

\begin{table*}[h]
    \centering
    \begin{adjustbox}{max width=0.8\textwidth}
    \begin{tabular}{ccccccccccc}
        \toprule
        & \textbf{Edits} & \multicolumn{2}{c}{\textbf{Bilingual 3k}} & \multicolumn{2}{c}{\textbf{Multilingual 3k}} & \multicolumn{2}{c}{\textbf{Bilingual 100}} & \multicolumn{2}{c}{\textbf{Multilingual 100}} \\
        \cmidrule(lr){3-4} \cmidrule(lr){5-6} \cmidrule(lr){7-8} \cmidrule(lr){9-10}
        & & \textbf{Acc} & \textbf{Hop-Acc} & \textbf{Acc} & \textbf{Hop-Acc} & \textbf{Acc} & \textbf{Hop-Acc} & \textbf{Acc} & \textbf{Hop-Acc} \\
        \midrule
        \multirow{8}{*}{\begin{turn}{90}\pokemqacl{}\end{turn}} & en & 31.5 & 23.3 & 13.1 & 5.4 & 41.8 & 31.8 & 27.7 & 12.6 \\
& de & 16.8 & 9.2 & 11.8 & 3.4 & 24.1 & 13.5 & 23.8 & 9.3 \\
& es & 18.5 & 8.9 & 10.8 & 2.9 & 25.4 & 12.1 & 22.0 & 7.2 \\
& hi & 7.0 & 0.1 & 9.8 & 1.1 & 12.7 & 0.8 & 14.7 & 2.7 \\
& sw & 11.8 & 5.7 & 11.9 & 2.3 & 14.9 & 8.2 & 21.9 & 5.0 \\
& bn & 7.0 & 0.2 & 8.0 & 0.5 & 14.0 & 0.5 & 12.0 & 1.6 \\
& ru & 8.0 & 0.6 & 10.7 & 1.4 & 17.4 & 2.9 & 18.6 & 5.0 \\
& zh & 8.4 & 0.5 & 9.1 & 1.2 & 15.0 & 2.4 & 16.7 & 3.5 \\
        \midrule
        & Average & 13.6 & 6.1 & 10.6 & 2.3 & 20.7 & 9.0 & \textbf{19.7}& 5.9 \\

        \midrule
        \multirow{8}{*}{\begin{turn}{90}\method{}\end{turn}} & en & 27.8 & 21.0 & 23.6 & 17.1 & 41.5 & 31.9 & 37.3 & 28.3 \\
& de & 23.5 & 13.7 & 19.7 & 12.1 & 29.5 & 18.6 & 26.4 & 17.4 \\
& es & 20.0 & 10.6 & 8.4 & 8.4 & 27.8 & 16.2 & 23.6 & 13.0 \\
& hi & 9.6 & 3.3 & 10.3 & 3.3 & 13.4 & 5.8 & 10.8 & 4.2 \\
& sw & 15.5 & 9.1 & 14.8 & 7.7 & 21.3 & 13.6 & 20.1 & 11.7 \\
& bn & 7.2 & 2.2 & 6.9 & 1.7 & 7.9 & 2.3 & 7.3 & 2.1 \\
& ru & 10.0 & 4.4 & 12.0 & 5.2 & 17.7 & 9.4 & 15.8 & 8.0 \\
& zh & 7.6 & 1.4 & 9.9 & 3.4 & 12.1 & 3.7 & 12.1 & 4.3 \\
        \midrule
        & Average & \textbf{15.1} & \textbf{8.2 }& \textbf{13.2} & \textbf{7.3 }& \textbf{21.4} & \textbf{12.7} & 19.2 & \textbf{11.1} \\
        \bottomrule
    \end{tabular}
    \end{adjustbox}
    \caption{Performance of \pokemqacl{} and \method{} by Language and Number of Edits on the \dataset{}-CF Dataset Using \llama{}-7B Backbone: Bilingual and Multilingual Training of the Retriever with All and 100 Edits.}
    \label{tab:llama-cf}
\end{table*}

\begin{table*}[h]
    \centering
    \begin{adjustbox}{max width=0.8\textwidth}
    \begin{tabular}{ccccccccccc}
        \toprule
        & \textbf{Edits} & \multicolumn{2}{c}{\textbf{Bilingual 1.8k}} & \multicolumn{2}{c}{\textbf{Multilingual 1.8k}} & \multicolumn{2}{c}{\textbf{Bilingual 100}} & \multicolumn{2}{c}{\textbf{Multilingual 100}} \\
        \cmidrule(lr){3-4} \cmidrule(lr){5-6} \cmidrule(lr){7-8} \cmidrule(lr){9-10}
        & & \textbf{Acc} & \textbf{Hop-Acc} & \textbf{Acc} & \textbf{Hop-Acc} & \textbf{Acc} & \textbf{Hop-Acc} & \textbf{Acc} & \textbf{Hop-Acc} \\
        \midrule
        \multirow{8}{*}{\begin{turn}{90}\pokemqacl{}\end{turn}} & en & 73.1 & 58.1 & 25.6 & 16.6 & 73.4 & 58.2 & 30.7 & 19.8 \\
& de & 44.0 & 33.6 & 11.6 & 7.8 & 63.8 & 51.6 & 15.0 & 10.7 \\
& es & 52.9 & 38.5 & 11.6 & 5.7 & 63.3 & 47.1 & 18.6 & 9.2 \\
& hi & 10.3 & 3.2 & 8.0 & 3.9 & 12.7 & 3.9 & 10.5 & 4.6 \\
& sw & 45.4 & 33.8 & 13.5 & 4.7 & 47.6 & 35.0 & 16.3 & 6.8 \\
& bn & 5.6 & 1.0 & 5.0 & 2.1 & 7.0 & 1.6 & 7.3 & 3.3 \\
& ru & 10.5 & 5.1 & 8.7 & 3.6 & 13.4 & 7.2 & 12.2 & 6.2 \\
& zh & 4.1 & 1.9 & 5.1 & 2.1 & 6.4 & 3.3 & 6.2 & 2.4 \\
        \midrule
        & Average & 30.7 & 21.9 & 11.1 & 5.8 & 36.0 & 26.0 & 14.6 & 7.8 \\ 

        \midrule
        \multirow{8}{*}{\begin{turn}{90}\method{}\end{turn}} & en & 71.8 & 57.9 & 71.5 & 57.2 & 72.1 & 58.1 & 72.0 & 57.5 \\
& de & 63.2 & 50.4 & 59.6 & 48.1 & 63.5 & 50.5 & 62.2 & 50.1 \\
& es & 57.9 & 45.0 & 51.6 & 40.0 & 58.0 & 45.1 & 52.7 & 40.8 \\
& hi & 33.2 & 19.0 & 25.4 & 15.0 & 34.9 & 20.1 & 27.9 & 16.2 \\
& sw & 43.1 & 33.1 & 45.3 & 33.7 & 44.0 & 33.6 & 46.7 & 34.6 \\
& bn & 10.3 & 5.8 & 7.8 & 4.6 & 10.5 & 5.8 & 9.6 & 5.2 \\
& ru & 58.5 & 37.2 & 30.3 & 18.6 & 62.4 & 40.5 & 34.3 & 21.1 \\
& zh & 40.5 & 29.0 & 33.7 & 22.8 & 42.0 & 30.1 & 35.0 & 23.6 \\
        \midrule
        & Average & \textbf{47.3} & \textbf{34.7} & \textbf{40.6} & \textbf{30.0} & \textbf{48.4} & \textbf{35.5} & \textbf{42.6} & \textbf{31.1} \\
        \bottomrule
    \end{tabular}
    \end{adjustbox}
    \caption{Performance of \pokemqacl{} and \method{} by Language and Number of Edits on the \dataset{}-T Dataset Using \llama-7B Backbone: Bilingual and Multilingual Training of the Retriever with All and 100 Edits.}
    \label{tab:llama-t}
\end{table*}

\begin{table*}[h]
    \centering
    \begin{adjustbox}{max width=0.8\textwidth}
    \begin{tabular}{ccccccccccc}
        \toprule
        & \textbf{Edits} & \multicolumn{2}{c}{\textbf{Bilingual 3k}} & \multicolumn{2}{c}{\textbf{Multilingual 3k}} & \multicolumn{2}{c}{\textbf{Bilingual 100}} & \multicolumn{2}{c}{\textbf{Multilingual 100}} \\
        \cmidrule(lr){3-4} \cmidrule(lr){5-6} \cmidrule(lr){7-8} \cmidrule(lr){9-10}
        & & \textbf{Acc} & \textbf{Hop-Acc} & \textbf{Acc} & \textbf{Hop-Acc} & \textbf{Acc} & \textbf{Hop-Acc} & \textbf{Acc} & \textbf{Hop-Acc} \\
        \midrule
        \multirow{8}{*}{\begin{turn}{90}\pokemqacl{}\end{turn}} & en & 28.6 & 21.8 & 13.5 & 5.4 & 37.5 & 29.5 & 25.5 & 13.0 \\
& de & 13.6 & 7.5 & 11.2 & 3.3 & 21.8 & 12.4 & 21.5 & 8.9 \\
& es & 18.2 & 9.5 & 10.5 & 2.7 & 23.1 & 12.7 & 19.6 & 7.2 \\
& hi & 6.8 & 0.2 & 7.9 & 0.8 & 11.9 & 0.7 & 13.3 & 2.0 \\
& sw & 11.4 & 6.3 & 10.3 & 2.5 & 14.5 & 8.3 & 17.5 & 5.3 \\
& bn & 6.1 & 0.2 & 6.2 & 0.4 & 13.4 & 0.3 & 9.7 & 1.0 \\
& ru & 7.4 & 0.6 & 7.8 & 1.0 & 14.4 & 2.6 & 16.1 & 4.2 \\
& zh & 8.0 & 0.3 & 8.7 & 0.7 & 13.3 & 2.0 & 15.0 & 2.6 \\
        \midrule
        & Average & 12.5 & 5.8 & 9.5 & 2.1 & 18.7 & 8.6 & 17.3 & 5.5 \\

        \midrule
        \multirow{8}{*}{\begin{turn}{90}\method{}\end{turn}} & en & 27.5 & 21.4 & 22.7 & 17.7 & 38.5 & 31.0 & 36.0 & 28.1 \\
& de & 19.6 & 12.8 & 17.5 & 12.0 & 27.2 & 17.8 & 25.9 & 17.6 \\
& es & 19.3 & 11.9 & 15.5 & 8.7 & 25.8 & 16.6 & 22.4 & 13.5 \\
& hi & 8.5 & 2.7 & 8.2 & 02.2 & 12.2 & 4.6 & 9.7 & 3.2 \\
& sw & 13.0 & 8.2 & 12.6 & 7.7 & 19.5 & 12.3 & 19.2 & 11.7 \\
& bn & 5.5 & 1.2 & 5.9 & 1.4 & 5.9 & 1.1 & 5.8 & 1.2 \\
& ru & 8.6 & 3.6 & 10.0 & 3.8 & 15.5 & 7.0 & 14.0 & 6.5 \\
& zh & 7.2 & 1.7 & 8.8 & 2.9 & 11.3 & 2.9 & 11.5 & 3.5 \\
        \midrule
        & Average & \textbf{13.6} & \textbf{7.9 }& \textbf{12.7} & \textbf{7.1 }& \textbf{19.5} & \textbf{11.7} & \textbf{18.1} & \textbf{10.7} \\
        \bottomrule
    \end{tabular}
    \end{adjustbox}
    \caption{Performance of \pokemqacl{} and \method{} by Language and Number of Edits on the \dataset{}-CF Dataset Using \vicuna-7B Backbone: Bilingual and Multilingual Training of the Retriever with All and 100 Edits.}
    \label{tab:vicuna-cf}
\end{table*}

\begin{table*}[h]
    \centering
    \begin{adjustbox}{max width=0.8\textwidth}
    \begin{tabular}{ccccccccccc}
        \toprule
        & \textbf{Edits} & \multicolumn{2}{c}{\textbf{Bilingual 1.8k}} & \multicolumn{2}{c}{\textbf{Multilingual 1.8k}} & \multicolumn{2}{c}{\textbf{Bilingual 100}} & \multicolumn{2}{c}{\textbf{Multilingual 100}} \\
        \cmidrule(lr){3-4} \cmidrule(lr){5-6} \cmidrule(lr){7-8} \cmidrule(lr){9-10}
        & & \textbf{Acc} & \textbf{Hop-Acc} & \textbf{Acc} & \textbf{Hop-Acc} & \textbf{Acc} & \textbf{Hop-Acc} & \textbf{Acc} & \textbf{Hop-Acc} \\
        \midrule
        \multirow{8}{*}{\begin{turn}{90}\pokemqacl{}\end{turn}} & en & 68.5 & 56.4 & 22.6 & 15.7 & 68.6 & 56.6 & 27.0 & 18.5 \\
& de & 59.1 & 47.5 & 10.3 & 7.2 & 59.4 & 47.7 & 13.6 & 9.6 \\
& es & 59.5 & 50.0 & 11.3 & 6.8 & 60.1 & 50.1 & 16.8 & 11.0 \\
& hi & 11.4 & 5.5 & 6.8 & 4.1 & 13.5 & 5.9 & 10.9 & 5.8 \\
& sw & 49.1 & 39.3 & 12.4 & 4.8 & 49.7 & 39.9 & 13.9 & 7.5 \\
& bn & 6.5 & 1.3 & 7.9 & 4.5 & 7.7 & 2.1 & 8.1 & 4.5 \\
& ru & 8.0 & 6.3 & 8.1 & 5.1 & 10.4 & 8.4 & 10.2 & 6.3 \\
& zh & 11.4 & 6.6 & 8.8 & 4.8 & 12.4 & 7.1 & 9.4 & 4.8 \\
        \midrule
        & Average & 34.2 & 26.6 & 11.0 & 6.6 &  35.2 & 27.2 & 13.7 & 8.5 \\ 

        \midrule
        \multirow{8}{*}{\begin{turn}{90}\method{}\end{turn}} & en & 69.0 & 57.3 & 68.0 & 56.5 & 69.2 & 57.5 & 68.8 & 57.0 \\
& de & 60.9 & 48.7 & 52.1 & 41.7 & 61.3 & 49.0 & 54.5 & 43.8 \\
& es & 56.9 & 47.3 & 49.6 & 41.8 & 57.0 & 47.3 & 51.0 & 42.7 \\
& hi & 23.4 & 14.8 & 24.1 & 16.9 & 26.0 & 16.9 & 27.1 & 19.0 \\
& sw & 44.4 & 36.6 & 47.3 & 39.9 & 45.3 & 37.5 & 48.7 & 41.0 \\
& bn & 11.3 & 08.0 & 11.4 & 08.5 & 11.1 & 08.0 & 13.2 & 09.3 \\
& ru & 51.9 & 40.5 & 26.4 & 20.7 & 55.5 & 44.3 & 28.9 & 22.9 \\
& zh & 32.5 & 24.5 & 24.7 & 19.0 & 34.5 & 26.3 & 27.1 & 19.0 \\
        \midrule
        & Average & \textbf{43.8} & \textbf{34.7} & \textbf{37.9} & \textbf{30.6} & \textbf{45.0} & \textbf{35.8} & \textbf{39.9} & \textbf{31.8}\\
        \bottomrule
    \end{tabular}
    \end{adjustbox}
    \caption{Performance of \pokemqacl{} and \method{} by Language and Number of Edits on the \dataset{}-T Dataset Using \vicuna-7B Backbone: Bilingual and Multilingual Training of the Retriever with All and 100 Edits.}
    \label{tab:vicuna-t}
\end{table*}

\end{document}